\documentclass[twoside]{article} % For LaTeX2e

\usepackage[accepted]{aistats2015}
\usepackage[colorlinks,linkcolor=blue]{hyperref}
\usepackage{url}
\usepackage[numbers]{natbib}
\usepackage{scabby}

\providecommand{\e}{\epsilon}
\providecommand{\La}{\Lambda}

\providecommand{\vt}{\tilde{v}}
\providecommand{\ut}{\tilde{u}}
\providecommand{\pit}{\tilde{\pi}}
\providecommand{\Vt}{\tilde{V}}
\providecommand{\Ut}{\tilde{U}}
\newcommand{\supr}[1]{^{(#1)}}

\providecommand{\sMh}{\widehat{\sM}}
\providecommand{\Mh}{\widehat{M}}
\providecommand{\Th}{\widehat{T}}

\providecommand{\pb}{\bar{p}}
\providecommand{\pib}{\bar{\pi}}
\providecommand{\Ub}{\bar{U}}
\providecommand{\Vb}{\bar{V}}
\providecommand{\Lb}{\bar{\Lambda}}

\begin{document}

\twocolumn[
%\title{Random Projections and Simultaneous Diagonalization: Improved Algorithms for Tensor Factorization}
% another idea (VK):
% PL: I like this title better since it's shorter
% Would be even better if it were more distinctive
%\aistatstitle{Tensor factorization via random projections and simultaneous matrix diagonalization}
%\aistatstitle{Project and Diagonalize: A Simple Approach to Tensor Factorization}

% PL: focus more on the reduction from tensors to matrix, which is the key part;
% previous title was too clunky
\aistatstitle{Tensor Factorization via Matrix Factorization}

%\aistatsauthor{ Anonymous Author 1 \And Anonymous Author 2 \And Anonymous Author 3 }
\aistatsauthor{ Volodymyr Kuleshov$^{*}$ \And Arun Tejasvi Chaganty$^{*}$ \And Percy Liang }
%\aistatsaddress{ Unknown Institution 1 \And Unknown Institution 2 \And Unknown Institution 3 }
 \aistatsaddress{ Department of Computer Science\\ Stanford University\\ Stanford, CA 94305}
]

\newcommand\blfootnote[1]{%
  \begingroup
  \renewcommand\thefootnote{}\footnote{#1}%
  \addtocounter{footnote}{-1}%
  \endgroup
}
\newcommand{\fix}{\marginpar{FIX}}
\newcommand{\new}{\marginpar{NEW}}

\begin{abstract}
%\todo{Tensor factorization vs. tensor decomposition?}
%\pl{Hm, Anandkumar and Kolda use 'tensor decomposition',
%but 'factorization' is derivationally more convenient -
%it's easier to say 'factorize a tensor', 'factors of a tensor'
%}

% PL: normally, we want to show broad applicability of a ML technique - mention other fields,
% but since this paper is a numerical methods paper - need to focus more on ML.
% PL: also, want to set a theoretical tone early on
%Multiple tasks in machine learning, neuroscience, and signal processing require factorizing a tensor.

%\vk{I like the new title, but do we have to do something extra to notify the committee that we changed it?}
Tensor factorization arises in many machine learning applications,
such as knowledge base modeling and parameter estimation in latent variable models.
%requiring a compact representation of multi-relational data --- for example, it
%has been the workhorse of recently developed consistent estimators for
%latent-variable models.
%
%\pl{say that tensor decomposition is actually difficult - more compelling than saying people haven't worked on tensors}
%\pl{also need to say much more about the challenges/weaknesses that plague previous methods (ALS, tensor power method)}
%\pl{unlike TPM, our method works for non-orthogonal, non-symmetric, no whitening (careful, we don't have proof)}
% PL: it's not like we're doing SVD - non-orthogonal simultaneous diagonalization is kind of sketchy.
%However, methods for factorizing tensors are not nearly as well developed as methods for factorizing matrices.
%
%Here, we propose a technique for reducing the problem of determining an undercomplete \pl{necessary?} canonical (CP) decomposition of a third-order tensor to that of simultaneously diagonalizing a set of matrices.
%However, compared to matrices, tensors cause more theoretical and algorithmic headaches.
%AC: I don't like this.
%However, factorizing a tensor is still not a fully solved problem, especially compared to the problem of diagonalizing matrices.
However, numerical methods for tensor factorization have not reached the
  level of maturity of matrix factorization methods.
In this paper, we propose a new algorithm for
  CP tensor factorization that uses random projections to reduce
  the problem to simultaneous matrix diagonalization.
Our method is conceptually simple and also applies to non-orthogonal and
  asymmetric tensors of arbitrary order.
We prove that a small number random projections essentially preserves
  the spectral information in the tensor, allowing us to
  remove the dependence on the eigengap that plagued earlier tensor-to-matrix reductions.
Experimentally, our method outperforms existing tensor factorization methods
  on both simulated data and two real datasets.
\end{abstract}

\section{Introduction}
\label{sec:intro}

% Define problem, give applications
Given a tensor $\Th \in \Re^{d \times d \times d}$ of the following form:
\begin{align}
  \label{eqn:cp}
\Th = \sum_{i=1}^k \pi_i a_i \otimes b_i \otimes c_i + \text{noise},
\end{align}
our goal is to estimate the factors $a_i, b_i, c_i \in \Re^{d}$ and factor weights $\pi \in \Re^k$.
In machine learning and statistics,
  this tensor $\Th$ typically represents
  higher-order relationships among variables,
  and we would like to uncover the salient factors that explain these relationships.
  %topics in a document, activation patterns of
  %neurons, or ways to compress an image.
This problem of \emph{tensor factorization} is an important problem rich with applications \citep{kolda2009tensor}:
  %statistical relational learning \citep{sutskever2009modelling,jenatton12latent},
  modeling knowledge bases \cite{nickel2011three},
  topic modeling \cite{anandkumar12lda},
  community detection \cite{anandkumar2013community},
  learning graphical models \cite{halpern2013unsupervised,chaganty2014graphical}.
The last three fall into a class of procedures based on the method of moments for latent-variable models,
  which are notable because
  they provide guarantees of consistent parameter estimation \cite{anandkumar13tensor}.

However, tensors, unlike matrices, are fraught with difficulties:
  identifiability is a delicate issue \citep{kruskal77three,desilva2008tensor,brachat2010symmetric},
  and computing \equationref{cp} is in general NP-hard \citep{hastad1990tensor,hillar2013tensor}.
% PL: made this description simpler
In this work, we propose a simple procedure to reduce the problem of factorizing tensors to that of factorizing matrices.
Specifically, we first project the tensor $\hat T$ onto a set of random vectors, producing a set of matrices.
Then we simultaneously diagonalize the matrices, producing an estimate of the factors of the original tensor.
We can optionally refine our estimate by running
the procedure using the estimated factors rather than random vectors.
Our approach applies to orthogonal, non-orthogonal and asymmetric 
  tensors of arbitrary order.
%But what are the advantages of such a reduction?

%% Our approach
%%\todo{State our assumption that $k \le d$.}
%In this work, we propose a new simple two-step approach to tensor factorization:
%first, we randomly project the tensor $\hat T$ to obtain a set of matrices,
%which can be simultaneously diagonalized to obtain initial estimates of the factors.
%Then, we project the tensor on these factors to obtain a new set of matrices,
%which can be simultaneously diagonalized again to yield the final estimates.
%% AC: Not sure
%We show that this algorithm has comparable accuracy to the reduction of
%  \citet{delathauwer2006decomposition} while being an more
%  efficient than the tensor power method \cite{anandkumar13tensor}\unneeded.

% AC: THis is a hack, placing here so there are not 2 footnotes on top of each other.
\blfootnote{* These authors contributed equally.}
% Reduction to simultaneous diagonalization
From a practical perspective,
this approach enables us to immediately leverage mature algorithms for matrix factorization.
Such algorithms often have readily available implementations that are numerically stable and highly
optimized.
In our experiments, we observed that they contribute to improvements in accuracy and speed over methods that deal directly with a tensor.
%(partly by operating only on $d^2$ and not $d^3$-dimensional objects).

From a theoretical perspective, we consider both {\em statistical} and 
{\em optimization} aspects of our method.
Most of our results pertain to the former:
we provide guarantees on the accuracy of a solution as a function of the noise $\e$ 
(this noise typically comes from the statistical estimation of $T$ from finite data)
that are comparable to those of existing methods (\tableref{method-shootout}).
Algorithms based on matrix diagonalization
have been previously criticized \cite{anandkumar13tensor} to be extremely sensitive to
noise due to a dependence on the smallest difference between eigenvalues (the
eigengap).
We show that this dependence can be entirely
  avoided using just $O(\log k)$ tensor projections chosen uniformly at random.
Furthermore, our guarantees are independent of the algorithm used for diagonalizing the projection matrices.

The {\em optimization} aspects of our method, on the other hand, depend on the choice of joint diagonalization subroutine.
Most subroutines
enjoy local quadratic convergence rates
\cite{yeredor2002non,ziehe2004fast,vollgraf2006quadratic}
and so does our method.
With sufficiently low noise, global convergence guarantees can be established for some joint diagonalization algorithms \cite{kuleshov2015simultaneous}. 
More importantly, local optima are not an issue for our method in practice, which is in sharp contrast to some other approaches, such as 
expectation maximization (EM).
%However we show
%empirically that they almost surely converge to the point at which our
%statistical results hold, and establish formal requirements on the
%diagonalization subroutine for that to be provably the case.

% Empirical results
Finally, we show that our method obtains accuracy improvements
  over alternating least squares and the tensor power method on several synthetic and real datasets.
On a community detection task, we obtain up to a 15\% reduction in error compared to a recently proposed approach \cite{anandkumar2013community},
and up to an 8\% reduction in error on a crowdsourcing task
\cite{zhang2014crowdsourcing}, matching or outperforming a state-of-the-art
EM-based estimator on three of the four datasets.

\paragraph{Notation}

Let $[n] = \{ 1, \dots, n \}$ denote the first $n$ positive integers.
Let $e_i$ be the indicator vector which is $1$ in component $i$ and $0$ in all other components.
We use $\otimes$ to denote the tensor product: if $u, v, w \in \Re^d$,
  then $u \otimes v \otimes w \in \Re^{d \times d \times d}$.\footnote{
  We will only consider third order tensors for the remainder of this paper,
though the approach naturally extends to tensors of arbitrary order.}
For a third order tensor $T \in \Re^{d \times d \times d}$ we define vector and matrix application as,
\begin{align*}
  T(x, y, z) &= \sum_{i=1}^d \sum_{j=1}^d \sum_{k=1}^{d} T_{ijk} x_i y_j z_k \\
  T(X, Y, Z)_{ijk}
  &= \sum_{l=1}^d \sum_{m=1}^d \sum_{n=1}^{d} T_{lmn} X_{li} Y_{mj} Z_{nk},
\end{align*}
for vectors $x, y, z \in \Re^{d}$ and matrices $X, Y, Z \in \Re^{d \times k}$.
The partial vector application (or projection) $T(I,I,w)$ of a vector
  $w \in \Re^d$ returns a $d \times d$ matrix:
$T(I, I, w)_{ij} = \sum_{k=1}^{d} T_{ijk} w_k$.

We define the CP decomposition of a tensor $T \in \mathbb R^{d \times d \times d}$ as
$T = \sum_{i=1}^k \pi_i a_i \otimes b_i \otimes c_i, $
for $a_i, b_i, c_i \in \mathbb R^d$. The rank of $T$ is said to be $k$.
When $a_i = b_i = c_i = u_i$ for all $i$, and the $u_i$'s are orthogonal,
we say $T$ has a symmetric orthogonal factorization,  $T = \sum_{i=1}^k \pi_i u_i^{\otimes 3}$.
Projecting a tensor $ T = \sum_{i=1}^k \pi_i a_i \otimes b_i \otimes c_i $ along $w$ produces a matrix $T(I,I,w) = \sum_{i=1}^k \pi_i (c_i^\top w) a_i \otimes b_i$. We use $\lambda_{i} = \pi_i (c_i^\top w)$ to refer to the factor weights (or eigenvalues in the orthogonal setting) of the projected matrix.
%\ac{We sometimes mention eigenvalues (it makes sense in the context) - adding the terms here.}
% We use $O(f(n))$ to denote a function $g(n)$ such that $\lim_{n \to\infty} g(n)/f(n) < \infty$.
%For a matrix $M \in \Re^{d\times d}$, let $\|M\|_F$ denote the Frobenius norm,
%% $\|M\|_F \eqdef (\sum_{i=1}^d \sum_{j=1}^d M_{ij}^2)^\half$
%let $\|M\|_\op \eqdef \max_{w} \frac{\|Mw\|_2}{\|w\|_2}$ denote the operator norm.
%, and let $\sigma_k(M)$ be the $k$-th largest singular value of $M$.
%Analogously, for a 3rd order tensor $T$, let
%$\|T\|_F$ be the Frobenius norm and let
%$\|T\|_F \eqdef (\sum_{i=1}^d \sum_{j=1}^d \sum_{k=1}^d T_{ijk}^2)^\half$
%$\|T\|_\op \eqdef \max_{w} \frac{\|T(I, w, w)\|_2}{\|w\|_2^2}$ be the operator norm.
% We say that a set of matrices $\{M_1, \cdots, M_L\}$, $M_l = \sum_{i=1}^d \lambda_{il} u_i v_i^\top$ has joint rank $k$
% if $\left| \{ i : \sum_{l=1}^L |\lambda_{il}| > 0\} \right| = k$.

For a vector of values $\pi \in \Re^k$, we use $\pi_{\min}$ and
$\pi_{\max}$ to denote the minimum and maximum absolute values
of the entries, respectively. Finally, we use $\delta_{ij}$ to
denote the indicator function, which equals $1$ when $i=j$ and $0$
otherwise.

% TODO(ARUN): Update the problem statement to work for arbitrary rank
% and stuff. Then defer to the 'extension' section.
% 
% \paragraph{Problem statement}
% 
% Let $T \in \Re^{d \times d \times d}$ be a rank-$k$ symmetric orthogonal tensor,
% $T = \sum_{i=1}^k \pi_i u_i \otimes u_i \otimes u_i$,
% where $u_i \in \Re^d$ are orthonormal vectors and $\pi \in \Re^k$ play
% the role of eigenvalues.
% Given a noisy observation of $T$, namely $\hat T \eqdef T + \e R$, where
%   $\e R$ is some additive noise where $\|R\|_\op = 1$,
%   we would like to recover $\pi$ and $v$ up to sign permutation.

\section{Background}
\label{sec:background}

In this section, we establish the context for tensor factorization,
method of moments for estimating latent-variable models, and simultaneous
matrix diagonalization.

%%%%%%%%%%%%%%%%%%%%%%%%%%%%%%%%%%%%%%%%%%%%%%%%%%%%%%%%%%%%
\subsection{Tensor factorization algorithms}
\label{sec:ten-fact-algos}

\begin{table}
\begin{center}
  \def\arraystretch{1.5}
  \begin{tabular}{c|c|c c c}
    Method	& $\mu < $ & $\|u_i - \ut_i\|_2 <$	& { Conv.} \\ \hline
TPM \cite{anandkumar13tensor}       
  & 0
& $\frac{\e}{\pi_{\min}}$
%& $\frac{k}{\pi_{\min}^2} \e$
  & G \\
Givens \cite{shalit2014coordinate}       
  & 0
& ?
  & G \\
ALS \cite{anandkumar2014guaranteed}          
& $\frac{\operatorname{polylog}(d)}{\sqrt{d}}$
& $\frac{\e}{\pi_{\min}} + \frac{\sqrt{k/d^{p-1}}}{\pi_{\min}}$
  & L \\
SD2  \cite{anandkumar12moments}                
  & 0
& $\frac{k^5 }{\pi_{\min} (\min_{i\neq j} |\pi_{i} - \pi_{j}|)} \e$
  & G \\
This paper 
  & 1
  & $ \frac{\|U\tinv\|_2^2}{(1-\mu^2) \pi_{\min}^2} \e $ 
  & L/G
\end{tabular}
\end{center}
\caption{Comparison of tensor factorization algorithms
(\sectionref{ten-fact-algos}).
For a tensor with noise $\epsilon$ (\equationref{cp})
and allowed incoherence $\mu$,
we show
an upper bound on the error in the recovered factors $\|u_i - \ut_i\|_2$ and
whether the convergence is (L)ocal or (G)lobal.
The factor weights $\pi$ are assumed to be normalized ($\|\pi\|_1 = 1$).
%\pl{this is too complicated; what are we trying to show?? can we just get rid of $\pi_{\min}$?}
%\pl{double check this all}
$\|U\tinv\|_2$ is the 2-norm of the inverse of factors $U^{-1}$.
Our method allows for arbitrary incoherence with a sensitivity to noise comparable to existing methods (\cite{anandkumar12moments,anandkumar13tensor,anandkumar2014guaranteed}), and with better empirical performance. 
%still being  \pl{fill in...}
In the orthogonal setting, our algorithm is globally convergent for sufficiently small $\e$.
%\pl{explain global/local for our method; isn't the situation the same for ALS}
}
\label{tbl:method-shootout}
\end{table}

Existing tensor factorization methods
  vary in their sensitivity to noise $\e$ in the tensor,
  their tolerance of non-orthogonality (as measured by the incoherence $\mu$)
  and in their convergence properties (\tableref{method-shootout}).
% Talk about orthogonal methods with guarantees
The robust tensor power method (TPM, \cite{anandkumar13tensor}) is a popular algorithm with
  theoretical guarantees on global convergence.
A recently-developed coordinate-descent
  method for orthogonal tensor factorization based on Givens rotations \cite{shalit2014coordinate}
  is empirically more robust than the TPM; however it is
  limited to the full-rank setting and lacks a sensitivity analysis.
% Non-orthogonal methods.
A further limitation of both methods is that they only work
  for symmetric orthogonal tensors.
  Asymmetric non-orthogonal tensors
  could be handled by preprocessing and whitening, but this can be
  a major source of errors in itself \citep{souloumiac2009joint}.
Alternating least squares (ALS) and other gradient-based methods \citep{comon2009tensor}
  are simple, popular, and apply to the non-orthogonal setting, but are
  known to easily get stuck in local optima \cite{delathauwer2006decomposition}.
\citet{anandkumar2014guaranteed} explicitly show both local and global
  convergence guarantees for a slight modification of the ALS procedure under certain assumptions on the tensor $\hat T$.

Finally, some authors have also proposed using simultaneous diagonalization for tensor factorization:
  \citet{delathauwer2006decomposition} proposed a reduction,
  but it requires forming a linear system of size $O(d^4)$ and is quite complex.
  %making the method prohibitively expensive.
\citet{anandkumar12moments} performed multiple random projections,
  but only diagonalized two at a time (SD2), leading to unstable results; the
  method also only applies to orthogonal factors.
\citet{anandkumar13tensor} briefly remarked that using all the projections at once was possible
  but did not pursue it.
In contrast, our method, has comparable bounds to the tensor power
  method in the orthogonal setting (conventionally $\|\pi\|_1 = 1$ is
  assumed), and the ALS method in the non-orthogonal setting.
Furthermore, in the non-orthogonal setting, our method works for arbitrary
  incoherence as long as the factors $U$ are non-singular.

  \vspace{-0.5em}
%%%%%%%%%%%%%%%%%%%%%%%%%%%%%%%%%%%%%%%%%%%%%%%%%%%%%%%%%%%%
\subsection{Parameter estimation in mixture models}
\label{sec:mom}

Tensor factorization can be used for parameter estimation for a wide range of
latent-variable models such as Gaussian mixture models, topic models, hidden
Markov models, etc. \cite{anandkumar13tensor}.
For illustrative purposes, we focus on the single topic model
\citep{anandkumar13tensor}, defined as follows:
%with $n$ documents with words coming from a vocabulary of $d$ words with $k$ topics.
For each of $n$ documents,
  draw a latent ``topic'' $h \in [k]$ with probability $\BP[h = i] = \pi_i$
  and three observed words $x_1, x_2, x_3 \in \{e_1, \dots, e_d\}$,
  which are conditionally independent given $h$ with $\BP[x_j = w \mid h = i] = u_{iw}$
  for each $j \in \{1, 2, 3\}$.
The parameter estimation task is
to output an estimate of the parameters $(\pi, \{u_i\}_{i=1}^k)$
  given $n$ documents $\{(x_1^{(i)},x_2^{(i)},x_3^{(i)}\}_{i=1}^n$
  (importantly, the topics are unobserved).

Traditional approaches typically use Expectation Maximization (EM) to optimize the marginal log-likelihood,
but this algorithm often gets stuck in local optima.  The method of moments approach is to cast
estimation as tensor factorization:
define the empirical tensor $\hat T = \frac1n \sum_{i=1}^n x_1^{(i)} \otimes x_2^{(i)} \otimes x_3^{(i)}$.
It can be shown that $\hat T = \sum_{i=1}^k \pi_i u_i \otimes u_i \otimes u_i + \epsilon R$ (a refinement of \equationref{cp}),
where $\epsilon R \in \Re^{d \times d \times d}$ is the statistical noise
which goes to zero as $n \to \infty$.
A tensor factorization scheme that asymptotically recovers estimates of
$(\pi,\{u_i\}_{i=1}^k)$ therefore provides a consistent estimator of the parameters.

%%%%%%%%%%%%%%%%%%%%%%%%%%%%%%%%%%%%%%%%%%%%%%%%%%%%%%%%%%%%
\subsection{Simultaneous diagonalization}
\label{sec:simdiag}

We now briefly review simultaneous matrix diagonalization,
the main technical driver in our approach.
%in the next section, we will show how tensor factorization can be reduced to this problem.
In simultaneous diagonalization,
we are given a set of symmetric matrices $M_1 ,\dots, M_L \in \mathbb R^{d \times d}$
(see \sectionref{asymmetric} for a reduction from the asymmetric case),
where each matrix can be expressed as
\begin{align}
M_l = U \Lambda_l U^\top + \epsilon R_l.
\end{align}
%\vk{I changed this so that it applies to low rank as well; Arun, could you double check please?}
The diagonal matrix $\Lambda_l \in \Re^{k \times k}$ and the noise $\epsilon R_l$ are individual
to each matrix, but the non-singular transform $U \in \Re^{d \times k}$ is common to all the matrices.
We also define the full-rank extensions,
\begin{align}
  \Ub &= \begin{bmatrix} U & U^\perp \end{bmatrix} &
  \Lb_l &= \begin{bmatrix} \Lambda_l & 0 \\ 0 & 0 \end{bmatrix},
\end{align}
where the columns of $U^\perp \in \Re^{d-k \times d}$ span the orthogonal subspace of $U$ and $\Lb_l \in \Re^{d \times d}$ has been appropriately padded with zeros. Note that $\Ub \Lb_l \Ub^\top = U \La_l U^\top$.

The goal is to find an invertible transform $V^{-1} \in \Re^{d \times d}$ such that each $V^{-1} M_l V^{-\top}$ is nearly diagonal. We refer to the $V^{-1}$ as {\em inverse factors}.
When $\epsilon = 0$, this problem admits a unique solution when there are at least two matrices
  \citep{afsari2008sensitivity}.
There are a number of objective functions for finding $V$
\cite{cardoso1996joint,yeredor2002non,afsari2006simple},
but in this paper, we focus on a popular one that penalizes off-diagonal terms:
%Simultaneous diagonalization algorithms determine the common $U$ by minimizing the objective
\begin{align}
\label{eqn:objective}
F(X)
\eqdef \sum_{l=1}^L \textrm{off}(X^{-1} M_l X^{-\top}),
\> \textrm{off}(A) = \sum_{i \neq j} A_{ij}^2.
%= \sum_{l=1}^L \sum_{i \neq j} (V^{-1} M_l V^{-\top})_{ij}^2.
\end{align}
%When $\e = 0$ and $L > 1$, it can be shown that $V=U$ is the unique minimizer of this objective function under a simple condition on a parameter called the modulus of unity \cite{afsari2006simple}. 
An important setting of this problem, which we refer to as the \emph{orthogonal case},
is when we know the true factors $U$ to be orthogonal. In this case we
constrain our optimization variable $X$ to be orthogonal as well, i.e.
$X^{-1} = X^\top$. 
%\pl{Not clear how $U$ and $V$ relate}
% When $k < d$, $V^{-1}$ is a pseudo-inverse. \vk{Right?}

%To obtain intuition, take $\epsilon = 0$.
%Then, the columns of $U$ are the common eigenvectors of each $M_l$,
%  and the optimal solution of \equationref{objective} is obtained by setting $V = U^\top$,
%  which makes $F(V) = 0$.
%Of course, in this case, 
In principle, we could just diagonalize one of the matrices, say $M_1$
  (assuming its eigenvalues are distinct) to recover $U$.
However, when $\epsilon > 0$,
  this procedure is unreliable and
  simultaneous diagonalization greatly improves on robustness to noise,
  as we will witness in \sectionref{orthogonal}.

There exist several algorithms for optimizing $F(X)$.
In this paper, we will use the Jacobi method
\cite{bunse1993numerical,cardoso1996joint} for the orthogonal case and the
QRJ1D algorithm \cite{afsari2006simple} for the non-orthogonal case.
Both techniques are based on same idea of iteratively constructing $X^{-1}$ via a product of simple matrices
$X^{-1} = B_T \cdots B_2 B_1$, where at each iteration $t = 1, \dots, T$, we choose $B_t$ to minimize $F(X)$.
Typically, this can be done in closed form.
%\todo{Our analysis is not algorithm specific about the global optimum}
%Typically, this can be done in closed form---see Appendix~\ref{sec:simdiagDetails} for details.
%\section{Simultaneous diagonalization}
%\label{sec:simdiagDetails}
%Recall that $V^{-1} = B_T \cdots B_2 B_1$,
%where at each iteration $t = 1, \dots, T$, we choose $B_t$ to minimize $J(V)$.

The Jacobi algorithm for the orthogonal case is a simple adaptation of the
Jacobi method for diagonalizing a single matrix.
Each $B_t$ is chosen to be a {\em Givens} rotation \cite{bunse1993numerical} defined by two of the $d$ axes $i < j \in [d]$:
%$B_t = \left( \begin{smallmatrix} \cos\theta & \sin\theta \\ -\sin\theta & \cos\theta \end{smallmatrix} \right) $
$B_t = (\cos\theta) (\Delta_{ii} + \Delta_{jj}) + (\sin\theta) (\Delta_{ij} - \Delta_{ji})$ 
for some angle $\theta$, where
$\Delta_{ij}$ is a matrix that is $1$ in the $(i,j)$-th entry and $0$
elsewhere.
We sweep over all $i<j$,
compute the best angle $\theta$ in closed form using the formula proposed by \citet{cardoso1996joint}
to obtain $B_t$, and then update each $M_l$ by $B_t M_l B_t^\top$.
The above can be done in 
%$O(k d^2 L)$ 
$O(d^3 L)$ 
time per sweep.
%using a sorted variant of the Jacobi algorithm \vk{Citation needed}.%\citet{cardoso1996joint}.

For the non-orthogonal case, the QRJ1D algorithm is similar,
  except that $B_t$ is chosen to be either a lower or upper unit triangular matrix
  ($B_t = I + a \Delta_{ij}$ for some $a$ and $i \neq j$).
The optimal value of $a$ that minimizes $F(X)$ can also be computed in closed form
(see \cite{afsari2006simple} for details).
The running time per iteration is the same as before.

\vspace{-0.5em}
\section{Tensor factorization via simultaneous matrix diagonalization}
\label{sec:algorithm}
\vspace{-0.5em}

We now outline our algorithm for symmetric third order tensors. In
\sectionref{asymmetric}, we describe how to generalize our method to
arbitrary tensors.
%
%To better understand how to reduce tensor factorization to matrix
%factorization, consider a tensor with an orthogonal decomposition $T = \sum_i
%\pi_i u_i^{\otimes 3}$. 
Observe that the projection of $T = \sum_i
\pi_i u_i^{\otimes 3}$ along a vector $w$
is a matrix $T(I, I, w) = \sum_i \pi_i (w^\top u_i) u_i^{\otimes 2}$ that
preserves all the information about the factors $u_i$
(assuming the $\pi_i (w^\top u_i)$'s are distinct).
In principle
one can recover the $u_i$ through an eigendecomposition of $T(I, I, w)$.
However, this method is sensitive to
noise: the error $\|u_i - \ut_i\|_2$ of an estimated eigenvector $\ut_i$ depends
on the reciprocal of the smallest eigengap $\max_{j\neq i} 1 / |\lambda_i - \lambda_j|$ of the
projected matrix (recall that $\lambda_i = \pi_i (w^\top u_i)$), which can be large and lead to inaccurate estimates.

Instead, let us obtain the factorization of $T$
from projections along multiple vectors $w_1, w_2, \cdots, w_L$. The
projections produce matrices of the form $M_l = \sum_{i}  \lambda_{il}
u_i^{\otimes 2}$, with $\lambda_{il} = \pi_i w_l^\top u_i$; they have common
eigenvectors, and therefore can be simultaneously diagonalized.
As we will show later, joint  diagonalization is sensitive to the
measure
$\min_{i\neq j} \sum_{l=1}^L (\lambda_{il} - \lambda_{jl})^2 / \left( \sum_{l=1}^L (\lambda_{il} - \lambda_{jl})^2 \right)$,
which averages the minimum eigengap across the matrices $M_l$ (here, $\lambda_{il} = \pi_i (w_l^\top u_i)$).
%Furthermore, in the case of non-orthogonal simultaneous
%  diagonalizations, multiple projections are necessary in order to uniquely recover the correct diagonalizer.
%
% chosen in a way that results in a large {\em average} eigengap  $|\lambda_i - \lambda_j|$. 
% Observe that projecting $T$ along $w_1, ..., w_L$ produces matrices of the form $ \sum_{i}  \lambda_{ik} u_i^{\otimes 2}$, with $\lambda_{ik} = \pi_i w_k^\top u_i$; they have common eigenvectors, and therefore can be jointly diagonalized.
%In the non-orthogonal setting, the factorization is not unique without multiple projections.

A natural question to ask is along which vectors $(w_l)$ should we project? In
\sectionref{orthogonal} and \sectionref{non-ortho-analysis} we show
that (a) estimates of the inverse factors $(v_i)$ are a good choice (when the $(v_i)$ are approximately orthogonal, they are close to the factors $(u_i)$) 
 and
that (b) random vectors do almost as well.
This suggests a simple two-step method: 
(i) first, we find approximations of the tensor factors by simultaneously diagonalizing
a small number of random projections of the tensor; (ii) then we perform another round of simultaneous
diagonalization on projections along the inverse of these approximate factors. 
%The choice of simultaneous diagonalization algorithm depends on whether
  %or not we know that the tensor has a symmetric orthogonal decomposition.\reword
\algorithmref{joint} describes the approach.
Its running time is $O(k^2d^2s)$, where $s$ is the number of sweeps of the simultaneous diagonalization algorithm. 
%It's $O(d^3k^2)$ and $O(d^2k^2)$ for low noise vs. $O(d^4k)$ for the tensor power method.

% The accuracy of the above method depends on how we chose the $w_l$. As we will show below, the best $w_l$ should be estimates of the true $u_i$. To find such estimates, we introduce a two-stage method. First, it chooses random $w_l$ and produce initial estimates $\ut_i\supr{0}$ that will be only slightly more accurate than if we used a single projection. Then, at the second step, it projects $T$ on the $\ut_i\supr{0}$ to produce new estimates $\ut_i\supr{1}$ that are substantially accurate both in practice and in theory, as we will see below. The formal description of our method is in Algorithm \ref{algo:joint}.
% 
% \todo{How many random projections are required in each case?}
% 
\begin{algorithm}[t]
  \caption{Two-stage tensor factorization algorithm}
  \label{algo:joint}
  \begin{algorithmic}[1]
    \REQUIRE $\Th = T + \e R \in \Re^{d \times d \times d}$, where $T$ has a CP decomposition $T = \sum_{i=1}^k \pi_i u_i^{\otimes 3}$, $L_0 \geq 2$
    \ENSURE Estimates of factors, $\pit, \ut_1, \cdots, \ut_k$.
    \STATE Define $\mathcal M\supr{0} \gets \{\hat T(I, I, w_l)\}_{l=1}^{L_0}$ with $\{w_l\}_{l=1}^{L_0}$ are chosen uniformly from the unit sphere $\mathcal S^{d-1}$.
    \STATE Obtain factors $\{\ut_i\supr{0}\}_{i=1}^{k}$ and their inverse $\{\vt_i\supr{0}\}_{i=1}^k$ from the simultaneous diagonalization of $\sM\supr{0}$.
    \STATE Define $\mathcal M\supr{1} \gets \{\hat T(I, I, \vt_i\supr{0})\}_{i=1}^{k}$.
    \RETURN Factors $\{\ut_i\supr{1}\}_{i=1}^{k}$ and factor weights $\{\pit_i\}_{i=1}^k$ from simultaneously diagonalizing $\sM\supr{1}$.
  \end{algorithmic}
\end{algorithm}
% 
% The crucial subroutine of Algorithm \ref{algo:joint} is simultaneous diagonalization. Any algorithm can be used at that stage, which leads to several advantages.
% \begin{itemize}
% \item Both the orthogonal and the non-orthogonal case are handled by this method. In the non-orthogonal case, at least two projections are needed to guarantee a unique eigendecomposition.
% \item Because matrix factorization is better studied that tensor factorization, the resulting algorithms are more mature and often faster.
% \item Our theoretical guarantees hold for any algorithm that solves the simultaneous diagonalization problem. They are usually simple because they leverage existing analyses for joint diagonalization algorithms.
% \item As better simultaneous diagonalization algorithms become available, our method will improve.
% \end{itemize}

%\subimport{}{theory}
\section{Perturbation analysis for orthogonal tensor factorization}
\label{sec:orthogonal}

In this section, we will focus on the orthogonal setting, returning to
  non-orthogonal factors in \sectionref{non-ortho-analysis}.
For ease of exposition, we restrict ourselves to symmetric third-order
  orthogonal tensors: $T = \sum_{i=1}^k \pi_i u_i^{\otimes 3}$.
Here the inverse factors $(v_i)$ are equivalent to the factors $(u_i)$,
  and we do not distinguish between the two.
The proofs for this section can be found in \appendixref{ortho-proofs}.

Our sensitivity analysis builds on the perturbation analysis result for
the simultaneous diagonalization of matrices in
\citet{cardoso1994perturbation}.
\begin{lemma}[\citet{cardoso1994perturbation}]
  \label{lem:cardoso}
  Let $M_l = U \La_l U^\top + \e R_l$, $l \in [L]$, be matrices with common factors $U \in \mathbb R^{d \times k}$ and diagonal $\Lambda_l \in \mathbb R^{k \times k}$.
  %Let $M_l = \sum_{i=1}^k \lambda_{il} u_i u_i^\top$, $l=1,2,...,L$ be matrices of joint rank $k$.
  Let $\Ub \in \Re^{d \times d}$ be a full-rank extension of $U$ with columns
  $u_1, u_2, \dots, u_d$ and let $\Ut \in \Re^{d \times d}$ be
  the orthogonal minimizer
  of the joint diagonalization objective $F(\cdot)$. 
  Then, for all $u_j$, $j \in [k]$, there exists a column $\tilde u_j$ of $\Ut$ such that
  \begin{align}
    \|\ut_j - u_j\|_2 \le \e \sqrt{\sum_{i=1}^{d} E_{ij}^2} + o(\e),
  \end{align}
where $E \in \mathbb R^{d \times k}$ is
\begin{align}
  E_{ij} 
  &\eqdef \frac{\sum_{l=1}^L (\lambda_{il} - \lambda_{jl}) u_j^\top R_l u_i}
  {\sum_{l=1}^L (\lambda_{il} - \lambda_{jl})^2} \label{eqn:e-ij-bound-0}
\end{align}
when $i\neq j$ and $i \leq k$ or $j \leq k$. We define $E_{ij} = 0$ when $i = j$ and $\lambda_{il} = 0$ when $i > k$.
\end{lemma}

In the tensor factorization setting, we jointly diagonalize projections $\Mh_l$, $l=1,2,\dots,L$ of the noisy tensor $\Th$ along vectors $w_l$: $\Mh_l = \Th(I, I, w_l) = \sum_{i=1}^k \pi_i (w_l^\top u_i) u_i\tp{2} + \epsilon R(I, I, w_l)$, where $R_l \eqdef R(I, I, w_l)$ has unit operator
  norm.
%Note that the projection matrices have joint rank $k \le d$. 
Cardoso's lemma provides bounds on the accuracy of recovering the $u_i$ via joint diagonalization; in particular,
we can further rewrite \equationref{e-ij-bound-0} in the tensor setting as:
\begin{align}
  E_{ij} 
  &= \frac{\sum_{l=1}^L w_l^\top p_{ij} r_{ij}^\top w_l}
  {\sum_{l=1}^L w_l^\top p_{ij} p_{ij}^\top w_l}, \label{eqn:e-ij-bound-2}
\end{align}
where $p_{ij} \eqdef (\pi_{i} u_i - \pi_{j} u_j)$ and $r_{ij} \eqdef R(u_i, u_j, I)$.

%\equationref{e-ij-bound-1} tells us that the error in recovering
%$u_i$ should depend on an eigengap $\min_{i,j} \sum_{i=1}^L (\lambda_{il} - \lambda_{jl}) / \left( \sum_{i=1}^L (\lambda_{il} - \lambda_{jl})^2 \right) that is
%averaged across the $L$ projection matrices$
%instead of the minimum eigengap.
\equationref{e-ij-bound-2} tells us that we can control the magnitude of the $E_{ij}$ (and hence the error on recovering the $u_i$)
  through appropriate choice of the projections $(w_l)$.
Ideally, we would like to ensure that the projected eigengap, $\min_{i \neq j} w_l^\top p_{ij} = \min_{i \neq j} \left(\pi_i(w_l^\top u_i) - \pi_j(w_l^\top u_j) \right)$, is
bounded away from zero for at least one $M_l$ so that the denominator of \equationref{e-ij-bound-2} does not blow up.
%  \pl{how is eigengap defined?  say what minimum eigengap is and why? need to explain this more generally}
%  \vk{did I address Percy's comment?}

% One way of ensuring this is to project along the true eigenvectors $u_i$:
  % then at least one matrix will have a large eigengap and the error term
  % $E_{ij}$ will be bounded. Of course, the true $u_i$ are unknown, but we can
  % do almost as well by either using ``plug-in" estimates of the $u_i$, or by
  % using many random projections. We formalize these two claims in the
  % theorems below.

\paragraph{Random projections}

The first step of \algorithmref{joint} projects the tensor along random
directions. The form of \equationref{e-ij-bound-2} suggests that the error
terms, $E_{ij}$, should concentrate over several projections and we will show that
this is indeed the case. Consequently, the error terms will depend inversely on the mean of $w_l^\top p_{ij}$,
$\|p_{ij}\|_2^2 = \pi_i^2 + \pi_j^2 > \pi_{\min}^2$.
%, instead of the eigengap, $\pi_i - \pi_j$
%\pl{this is not the eigengap...}. \vk{can we remove the last sentence?}
Our final result is as follows:

%Of course, we do not have access to the eigenvectors or even
%approximations of them a priori. 
%However, it turns out that by choosing a small number of random projections, we can again ensure that
%the $E_{ij}$ are bounded with high probability such that there is again no dependence on the eigengap.
%We will now show that this is indeed the case. 
% First, we show that each element $(E_{ij})$ is bounded with high
% probability,
% \begin{lemma}[Concentration of diagonalization errors]
%   \label{lem:e-ij-conc-1}
% Let $w_1, \ldots, w_L$ be i.i.d.~random Gaussian vectors, $w_l \sim
%   \sN(0, I)$, and let $\sMh = \{\Mh_1, \ldots,
%   \Mh_L\}$ be constructed via projection of $\Th$ along $w_1, \ldots,
%   w_L$.
% Further, assume $L \ge 16\log(2\delta)^2$. 
% Then, with probability at least $1 - \delta$, 
% \begin{align*}
%     E_{ij} 
%         &\le
%         \frac{p_{ij}^\top r_{ij}}{\|p_{ij}\|_2^2} 
%             + \frac{10 \log(2/\delta)}{\sqrt L} \frac{\|r_{ij} \|_2}{\|p_{ij}\|_2}.
% \end{align*}
% \end{lemma}
% This leads to the following bound on the errors in the eigenvectors themselves.
\begin{theorem}[Tensor factorization with random projections]
  \label{thm:ortho-fact-random}
  Let $w_1, \ldots, w_L$ be i.i.d.~Gaussian vectors, $w_l \sim
    \sN(0, I)$, and let the matrices $\Mh_l \in \mathbb R^{d \times d}$ % = \{\Mh_1, \ldots, \Mh_L\}$
    be constructed via projection of $\Th$ along $w_1, \ldots,
    w_L$. 
%    Let $\ut_i$ be estimates of the $u_i$ found by orthogonal joint diagonalization of the $\Mh_l$.
  Let $\ut_i$ be estimates of the $u_i$ derived from the $\Mh_l$.
  Let $L \ge 16 \log(2 d (k-1)/\delta)^2$. Then, with probability at least $1-\delta$, for every $u_i$, there exists a $\ut_i$ such that
  \begin{align*}
    \| \ut_i - u_i \|_2 
      &\leq 
      \left( \frac{2\sqrt{2\|\pi\|_1 \pi_{\max}}}{\pi_{i}^2} + \frac{C(\delta)}{\pi_{i}} \right) \e
        + o(\e),
  \end{align*}
  where $C(\delta) \eqdef O\left(\log(kd)/\delta) \sqrt{\frac{d}{L}} \right)$.
\end{theorem}
The first of the above two terms is the fundamental error in estimating a noisy tensor $\hat T$;
the second term is due to the concentration of random projections and
can be made arbitrarily small by increasing $L$.

\paragraph{Plug-in projections}

The next step of our algorithm projects the tensor along the approximate factors from step 2.
Intuitively, if the $w_l$ are close to the eigenvectors $u_i$, then $w_l^\top p_{ij} = w_l^\top (\pi_i
u_i - \pi_j u_j) \approx \pi_i \delta_{il}$. Then for each $i \neq j$, there is some projection that ensures that $E_{ij}$ is bounded
and does not depend on the projected eigengap $\min_{i \neq j} \left(\pi(w_l^\top u_i) - \pi(w_l^\top u_j) \right)$.
%such that the eigengap between $u_i$ and $u_j$ is atleast $\pi_i$.
%With this choice, the error again no longer depends on the eigengap of the diagonalized matrices:
%\begin{lemma}[Diagonalization error with plug-in projections]
%  %\label{lem:e-ij-plug-in}
%Let $w_1, \ldots, w_k$ be approximations of $u_1, \ldots, u_k$: $\|w_l - u_l\|_2 \le \gamma$,
%  and let $\sMh = \{\Mh_1, \ldots, \Mh_L\}$ be constructed via
%  projection of $\Th$ along $w_1, \ldots, w_L$.
%If the set of matrices $\sMh$ is simultaneously diagonalized, then the
%  first-order error $E_{ij}$ is bounded from above:
%\begin{align*}
%  E_{ij} &\le \frac{ p_{ij}^\top r_{ij}}{\|p_{ij}\|^2} + O(\gamma).
%\end{align*}
%\end{lemma}
%When the approximation is on the same order as the perturbation, i.e.
%$\gamma \le C\e$, then we find that we recover the eigenvectors $u_i$
%well and with no dependence on the eigengap.
\begin{theorem}[Tensor factorization with plug-in projections]
  \label{thm:ortho-fact-plug-in}
  Let $w_1, \ldots, w_k$ be approximations of $u_1, \ldots, u_k$: $\|w_l - u_l\|_2 = O(\e)$,
    and let $\Mh \in \mathbb R^{d \times d}$ %= \{\Mh_1, \ldots, \Mh_k\}$ 
    be constructed via projection of $\Th$ along $w_1, \ldots, w_k$.
     Let $\ut_i$ be estimates of the $u_i$ derived from the $\Mh_l$.
  Then, for every $u_i$, there exists a $\ut_i$ such that
  \begin{align*}
    \| \ut_i - u_i \|_2 
    &\leq 
    \frac{2\sqrt{\|\pi\|_1 \pi_{\max}}}{\pi_{i}^2} \e 
     + o(\e).
  \end{align*}
\end{theorem}
%\pl{what is the dependence on $C$?
%better to write $\|w_l - u_l\|_2 = O(\epsilon)$ if we're not tracking it;
%point out it's interesting that can tolerate error on the order of $\epsilon$,
%but doesn't affect main term of error in $\tilde u_i$ (this is nice!)
%}

Note that \theoremref{ortho-fact-random} says that with $O(d)$
  random projections, we can recover the eigenvectors $u_i$ with almost the
  same precision as if we used approximate eigenvectors, with high
  probability.
Moreover, as $L \to \infty$, there is no gap between the precision of
  the two methods.
\theoremref{ortho-fact-plug-in} on the other hand suggests that we can
tolerate errors on the order of $O(\e)$ without significantly affecting
the error in recovering $\ut_i$.
In practice, we find that using the plug-in estimates allows us to improve
accuracy with fewer random projections.

\section{Perturbation analysis for non-orthogonal tensor factorization}
\label{sec:non-ortho-analysis}

We now extend our results to the case when the tensor $T$ has a non-orthogonal symmetric CP decomposition:
$T = \sum_{i=1}^k \pi_i u_i^{\otimes 3}$, where the $u_i$ are not orthogonal and $k \le d$.
We parameterize the non-orthogonality using incoherence: $\mu \eqdef \max_{i\neq j} u_i^\top u_j$ and the norm of the inverse factor $\|V^\top\|_2$ where $V \eqdef U\inv$. 
Compared to the orthogonal setting, our bounds reveal an
$O\left(\frac{\|V^\top\|_2^2}{1 - \mu^2}\right)$ dependence on
incoherence. Note that unlike previous work, our algorithm does not
require an explicit bound on $\mu$ (i.e. any $\mu < 1$ is sufficient),
as long as the factors $U$ are non-singular.
Proofs for this section are found in \appendixref{non-ortho-proofs}.

We base our analysis on the perturbation result by
\citet{afsari2008sensitivity}.

\begin{lemma}[\citet{afsari2008sensitivity}]
  \label{lem:afsari}
  Let $M_l = U \La_l U^\top + \e R_l$, $l \in [L]$, be matrices with common factors $U \in \mathbb R^{d \times k}$ and diagonal $\Lambda_l \in \mathbb R^{k \times k}$.
  %Let $M_l = \sum_{i=1}^k \lambda_{il} u_i u_i^\top$, $l=1,2,...,L$ be matrices of joint rank $k$.
% Let $U \in \mathbb R^{d \times k}$ be the matrix of (possibly non-orthogonal) common factors $u_i$, let $V$ be the minimizer
%   of the joint diagonalization objective $F(\cdot)$, and let $\Ut = V\pinv \in \Re^{d \times k}$.
  Let $\Ub \in \Re^{d \times d}$ be a full-rank extension of $U$ with columns
  $u_1, u_2, \dots, u_d$ and let $\Vb = \Ub\inv$, with rows $v_1, v_2, \dots, v_d$.
  Let $\Ut\in \Re^{d \times d}$ be
%  \pl{shouldn't this be $\tilde U$ to be consistent with the orthogonal case?}
  the minimizer
  of the joint diagonalization objective $F(\cdot)$ and let $\Vt = \Ut\inv$. 

  Then, for all $u_j$, $j \in [k]$, there exists a column $\tilde u_j$ of $\Ut$ such that
  \begin{align}
    \|\ut_j - u_j\|_2 \le \e \sqrt{\sum_{i=1}^d E_{ij}^2} + o(\e),
  \end{align}
where the entries of $E \in \mathbb R^{d \times k}$ are bounded by
\begin{align*}
  |E_{ij}|
  &\le \frac{1}{1 - \rho_{ij}^2} 
  \left(\frac{1}{\|\lambda_i\|^2_2} + \frac{1}{\|\lambda_j\|^2_2}\right) \\
  &\quad 
  \left( \left|\sum_{l=1}^L v_i^\top R_l v_j \lambda_{jl} \right| + \left|\sum_{l=1}^L v_i^\top R_l v_j \lambda_{il} \right| \right),
\end{align*}
when $i \neq j$ and $E_{ij} = 0$ when $i = j$  and $\lambda_{il} = 0$ when $i > k$.
Here $\lambda_i = (\lambda_{i1}, \lambda_{i2}, ..., \lambda_{iL}) \in \Re^L$ and
$\rho_{ij} = \frac{\lambda_{i}^\top \lambda_{j}}{\|\lambda_{i}\|_2
\|\lambda_{j}\|_2}$ is the modulus of uniqueness, a measure of
how ill-conditioned the problem is.
\end{lemma}

In the orthogonal case,
we had a dependence on the eigengap $\lambda_i - \lambda_j$.
Now the error crucially depends on the
modulus of uniqueness, $\rho_{ij}$. 
The non-orthogonal simultaneous diagonalization problem has a unique
solution iff $|\rho_{ij}| < 1$ for all $i \neq j$ \citep{afsari2008sensitivity}.  
In the orthogonal case, $\rho_{ij} = 0$.
It can be shown that $\rho_{ij}$ can once again be controlled by
  appropriately choosing the projections $(w_l)$. 

To get a handle on the difficulty of the problem, let us
assume that the vectors $u_i$ are incoherent: $u_i^\top u_j \le
\mu$ for all $i \neq j$. Intuitively, the problem is easy when $\mu \approx 0$
and hard when $\mu \approx 1$.

% Once again, we observe that having multiple projections should
% intuitively bound $\|\lambda_i\|_2$ away from zero and $\lambda_i^\top
% \lambda_j$ away from 1.

%Before we proceed, let us expose the structure induced because the
%matrices $(M_l)$ were generated by projections of a tensor, as well as
%introduce an incoherence assumption to measure how ``well behaved'' the
%problem is.
%Once again noting that $\lambda_{il} = \pi_i w_l^\top u_i$ and that $u_i^\top R_l u_j = r_{ij}^\top w_l$, we get the following expression,
%\begin{align}
%  |E_{ij}|
%  &\le \frac{1}{1 - \rho_{ij}^2} 
%  \left(\frac{1}{\|\lambda_i\|_2} + \frac{1}{\|\lambda_j\|_2}\right)
%  \left( |\pi_{j}| \left|\sum_{l=1}^L w_l^\top u_j r_{ij}^\top w_l \right| +
%    |\pi_{i}| \left|\sum_{l=1}^L w_l^\top u_i r_{ij}^\top w_l \right|
%    \right), \label{eqn:e-ij-non-ortho}
%\end{align}
%where $\|\lambda_i\|_2^2 = w_l^\top u_i u_i^\top w_l$, and $\rho_{ij}$ has the following expression,
%\begin{align}
%  \rho_{ij} 
%    &= \frac{\lambda_i^\top \lambda_j}{\|\lambda_i\|_2 \|\lambda_j\|_2 }
%    = \frac{\sum_{l=1}^L w_l^\top u_i u_j^\top w_l}{
%    \sqrt{ (\sum_{l=1}^L w_l^\top u_i u_i^\top w_l) (\sum_{l=1}^L w_l^\top u_i u_i^\top w_l)}}. \label{eqn:rho-ij}
%\end{align}

% Invoking black magic.
\paragraph{Random projections}
Intuitively, random projections are isotropic and hence we expect the
projections $\lambda_i$ and $\lambda_j$ to be nearly orthogonal to each
other. This allows us to show that $\rho_{ij} \le O(\mu)$, which matches our
intuitions on the difficulty of the problem. Our final result is the following:
%The following bound on the error of $\ut_i$ reflects this intuition.
\begin{theorem}[Non-orthogonal tensor factorization with random projections]
  \label{thm:non-ortho-random}
  Let $w_1, \ldots, w_L$ be i.i.d.~random Gaussian vectors, $w_l \sim
    \sN(0, I)$, and let the matrices $\Mh_l \in \mathbb R^{d \times d}$ % = \{\Mh_1, \ldots, \Mh_L\}$
    be constructed via projection of $\Th$ along $w_1, \ldots,
    w_L$. 
    Assume incoherence $\mu$ on $(u_i)$: $u_i^\top u_j \le \mu$.
  Let $L_0 \eqdef \left(\frac{50}{1 - \mu^2}\right)^2$ and 
  let $L \ge L_0 \log(15 d (k-1)/\delta)^2$. 
  Then, with probability at least $1-\delta$, for every $u_i$, there exists a $\tilde u_i$ such that
  {\small  
   \begin{align*}
    \|\ut_j - u_j\|_2 
      &\le O\left(
      \frac{\sqrt{\|\pi\|_1 \pi_{\max}}}{\pi_{\min}^2}  
      \frac{\|V^\top\|_2^2}{1-\mu^2} 
      \left(1 + C(\delta)\right)
      \right) \e + o(\e),
  \end{align*}
  }
  where $C(\delta) \eqdef \left(\log(kd/\delta) \sqrt{\frac{d}{L}} \right)$.
\end{theorem}

Once again, the error decomposes into a fundamental recovery error and
  a concentration term.
Note that the error is sensitive to the smallest factor weight,
$\pi_{\min}$. This dependence arises from the sensitivity of the
non-orthogonal factorization method to the $\lambda_i$ with the smallest
norm and is unavoidable.

\paragraph{Plug-in projections}
%In the orthogonal setting, the approximate factors ensured that
%there was at least one projection with non-zero eigengap. In particular,
%we got that that the vector $\lambda_i$ had one non-zero component with
%value $\pi_i$ and zeros elsewhere. 
%This ensures that the modulus of uniqueness, $\rho_{ij}$, is
%small, while $\|\lambda_i\|$ will be at least $|\pi_i|$.
When using plug-in estimates for the projections, two obvious
  choices arise: estimates of the columns of the factors, $(u_i)$, or
  the rows of the inverse, $(v_i)$. 
Using estimates of $(u_i)$ leads to $\rho_{ij} \le O(\mu)$, similar to what we saw
  with random projections. 
However, using estimates of $(v_i)$ ensures that the $\lambda_i$ are nearly orthogonal, resulting in $\rho_{ij} \approx 0$!
This leads to estimates that are less sensitive to the incoherence $\mu$.
% \pl{why not $d$ since $\mu \le 1/2d$?} (in \lemmaref{incoherence-projection}).
\begin{theorem}[Non-orthogonal tensor factorization with plug-in projections]
  \label{thm:non-ortho-plug-in}
  Let $w_1, \ldots, w_k$ be approximations of $v_1, \ldots, v_k$: $\|w_l - v_l\|_2 \le O(\e)$,
    and let the matrices $\Mh_l \in \mathbb R^{d \times d}$
    be constructed via projection of $\Th$ along $w_1, \ldots, w_k$.
%  Let $k \eqdef \left(\frac{(k-2)\mu + 2}{(k-1)\mu^2 + 1}\right)^2$.
  Also assume that the $u_i$ are incoherent: $u_i^\top u_j \le \mu$ when $j \neq i$.
   Then, for every $u_j$, there exists a $\tilde u_j$ such that
   \begin{align*}
    \|\ut_j - u_j\|_2 
      &\le O\left(
      %\frac{1 + \sqrt{\mu d}}{1 - k \mu^2} 
      \frac{\sqrt{\|\pi\|_1 \pi_{\max}}}{\pi_{\min}^2}  
      \|V^\top\|_2^3
      \right) \e + o(\e).
  \end{align*}
\end{theorem}

% \pl{need more intuition; how to think of $\mu_0$?  Can't we just upper bound $\mu_0$ by a constant?
%   that would be lovely;
%   we have now $\mu\sqrt{d}$ rather than $\sqrt{\mu d}$;
% is this expected?
% }

% \paragraph{Convergence properties.} As in the previous section, the above bounds hold in the vicinity of the perturbed global optimum. Since computing a CP-decomposition is NP-hard, strong formal guarantees may be difficult to establish in the non-orthogonal setting. However, our empirical experiments (Section \ref{sec:convergence}) still indicate that local optima are not an issue for small enough noise and for a bounded incoherence.

\section{Asymmetric and higher-order tensors}
\label{sec:asymmetric}

%In \sectionref{algorithm}, we presented \algorithmref{joint} to solve
%the tensor factorization problem for symmetric third-order tensors with
%orthogonal and non-orthogonal factors. 
In this section, we present
simple extensions to the algorithm to asymmetric and higher order tensors.

\paragraph{Asymmetric tensors}
We use a reduction to handle asymmetric tensors.
%used in SVD algorithms. 
Observe that the $l$-th projection $M_l$ of an
asymmetric tensor has the form $M_l = \sum_i \lambda_i u_{il} v_{il}^\top
= U \Lambda_l V^\top$, for some diagonal (not necessarily positive) matrix $\Lambda_l$ and
common $U,V$, not necessarily orthogonal. For each $M_l$, define another
matrix $ N_l = \left( \begin{smallmatrix}
    0 & M_l^\top \\
    M_l & 0
  \end{smallmatrix} \right)$ and observe that 
\begin{align*}
  \begin{bmatrix}
    0 & M_l^\top \\
    M_l & 0
  \end{bmatrix}
  &= 
  \frac{1}{2}
  \begin{bmatrix}
    V & V \\
    U & -U
  \end{bmatrix}
  \begin{bmatrix}
    \Lambda_l & 0 \\
    0 & -\Lambda_l
  \end{bmatrix}
  \begin{bmatrix}
    V & V \\
    U & -U \\
  \end{bmatrix}^\top.
\end{align*}
The $(N_l)$ are symmetric matrices with common (in general,
non-orthogonal) factors. Therefore, they can be jointly diagonalized and from their components, we can recover the components of the $(M_l)$. 
%This reduces the asymmetric case to a symmetric one. 
This reduction does not change the modulus of uniqueness of the problem: the
factor weights remain unchanged. 

\paragraph{Higher order tensors}
%\pl{cut this section if we don't have space since we don't actually use this; more important to explain other
%parts of the paper well}
Finally, if we have a higher order (say fourth order) tensor $T = \sum_{i} \pi_i a_i \otimes b_i \otimes c_i \otimes d_i$ then we can first determine the $a_i, b_i$ by projecting into matrices $T(I, I, w, u) = \sum_{i} \pi (w^\top c_i) (u^\top  d_i) a_i \otimes b_i$, and then determine the $c_i, d_i$ by projecting along the first two components.
Our bounds only depend on the dimension of the matrices being simultaneously diagonalized, and thus this reduction does not introduce additional error.
Intuitively, we should expect that additional modes of a tensor should provide more information and thus help estimation, not hurt it.
However, note that as the tensor order increases,
the noise in the tensor will presumably increase as well.
%tensors to within an $\e$ factor requires exponentially more samples.

%advantages: non-orthogonal case, noise sensitivity/guarantees, plug-in algorithm, simple analysis, speed, future improvements

\section{Convergence properties.}

% Study properties
%When do these algorithms work?
%\equationref{objective} is after all non-convex.
%\pl{fact check this}
The convergence of our algorithm depends on the choice of joint diagonalization subroutine.
Theoretically, the Jacobi method, the QRJ1D algorithm, and other algorithms are
guaranteed to converge to a local minimum at a quadratic rate
\cite{bunse1993numerical,ziehe2004fast,yeredor2004approximate}.
The question of global convergence is currently open
\cite{delathauwer2001independent, cardoso1996joint}.
%In practice, however, the Jacobi-style methods we adopt are well-known to behave
%as if they global convergence
Empirically though, these algorithms have been found in the literature to
converge reliably to global minima
\cite{bunse1993numerical,cardoso1996joint,delathauwer2001independent}
and to corroborate this claim, we conducted a series of experiments \cite{kuleshov2015simultaneous}.

We first examined convergence to global minima in the orthogonal setting. In 1000 trials of the Jacobi algorithm
on random sets of matrices for various $\e$ and $d=L=15$
%(\sectionref{convergence}) 
, we found that the objective values formed a Gaussian
distribution around $\e$ (the best accuracy that can be achieved). 
%\ac{We need to refer to arXiv here too?}
%Furthermore, setting $\e=10^{-4}, 10^{-3}, 10^{-2}$ resulted in the mean being multiplied by 10 each time, and the standard deviation was unchanged. 
Then, on each of our real crowdsourcing datasets, we ran our algorithm from 1000 random starting points; 
in every case, the algorithm converged to
the same solution (unlike EM). This suggests that our diagonalization
algorithm is not sensitive to local optima.
To complement this empirical evidence, we also established
that the Jacobi algorithm will converge to the global minimum when $\e$ is sufficiently small
and when the algorithm is initialized with the eigendecomposition of a single projection matrix  \cite{kuleshov2015simultaneous}.
%\ac{Include a reference to arXiv about the conditions under which Jacobi converges.}

We also performed similar experiments in the non-orthogonal setting using the QRJ1D algorithm. Unlike Jacobi, QRJ1D suffers from local optima, which is expected since the general CP decomposition problem is NP-hard.
However, local optima appear to only affect matrices with bad incoherence values, and in several real world experiments (see below), non-orthogonal methods fared better their orthogonal counterparts.

\section{Experiments}
\label{sec:evaluation}

In the orthogonal setting,
we compare our algorithms (\textbf{OJD0}, which uses random projections, and \textbf{OJD1} which uses with plug-in)
with the tensor power method (\textbf{TPM}), alternating least squares (\textbf{ALS}), and with
the method of de \textbf{Lathauwer} \cite{delathauwer2006decomposition}.
In the non-orthogonal setting, we compare de \textbf{Lathauwer},
alternating least squares (\textbf{ALS}), non-linear least squares (\textbf{NLS}),
and our non-orthogonal methods (\textbf{NOJD0} and \textbf{NOJD1}).

\begin{figure}
\begin{center}
%\framebox[4.0in]{$\;$}
\includegraphics[width=8cm]{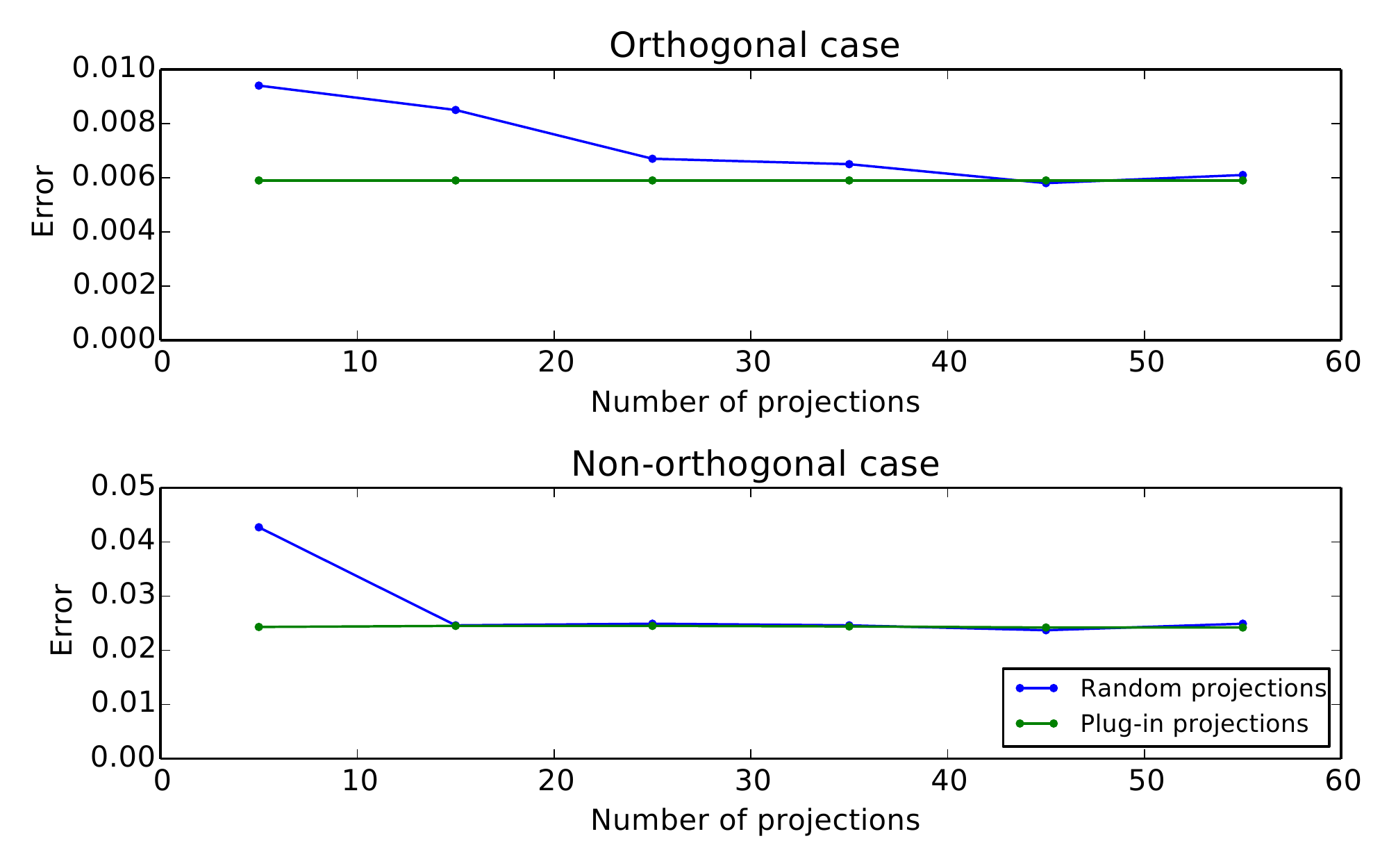}
\end{center}
\caption{Comparing random vs. plug-in projections ($d=k=10$, $\e_\textrm{ortho} =
0.05$, $\e_\textrm{nonortho} = 0.01$)
}\label{fig:proj-comparison}
\end{figure}

\paragraph{Random versus plug-in projections}
%We first evaluate the utility of plug-in over random projections.
We generated
random tensors $T = \sum_{i=1}^k \pi u_i^{\otimes 3} + \e R$ with Gaussian entries in
$\pi, R$ and $u_i$
distributed uniformly in the sphere $\mathcal S^{d-1}$.
%In the orthogonal setting, the $u_i$'s are constrained to be orthogonal.
In Figure \ref{fig:proj-comparison}, we plot the error $\sum_{i=1}^k \frac{1}{k} \| u_i - \ut_i \|_2$ (averaged over 1000 trials) of
using $L$ random projections (blue line), versus using $L$ random projections
followed by plug-in (green line). The accuracy of random projections tends to a limit that is immediately achieved by the plug-in projections, as predicted by our theory.
%
%In both the orthogonal and non-orthogonal settings,
%we see that plug-in estimates random projection error
%tends to a limit that is achieved much earlier by plug-in.
In the orthogonal setting,
plug-in reduces the total number of projected
matrices $L$ required to achieve the limiting error by three-fold (20 vs. 60 when $d=10$).
In the non-orthogonal setting, the difference between the
two regimes is much smaller.

\paragraph{Synthetic accuracy experiments}

We generated random tensors for various $d, k, \e$ using the same procedure as above. 
%let $d=25, 50, 100$ and in each case consider two
%regimes: undercomplete tensors with $k=0.2d$ and full rank tensors, $k=d$. 
We vary $\e$ and report the average error $\sum_{i=1}^k \frac{1}{k} \| u_i - \ut_i \|_2$ across 50 trials.
%

\begin{figure}
\begin{center}
\includegraphics[width=8cm]{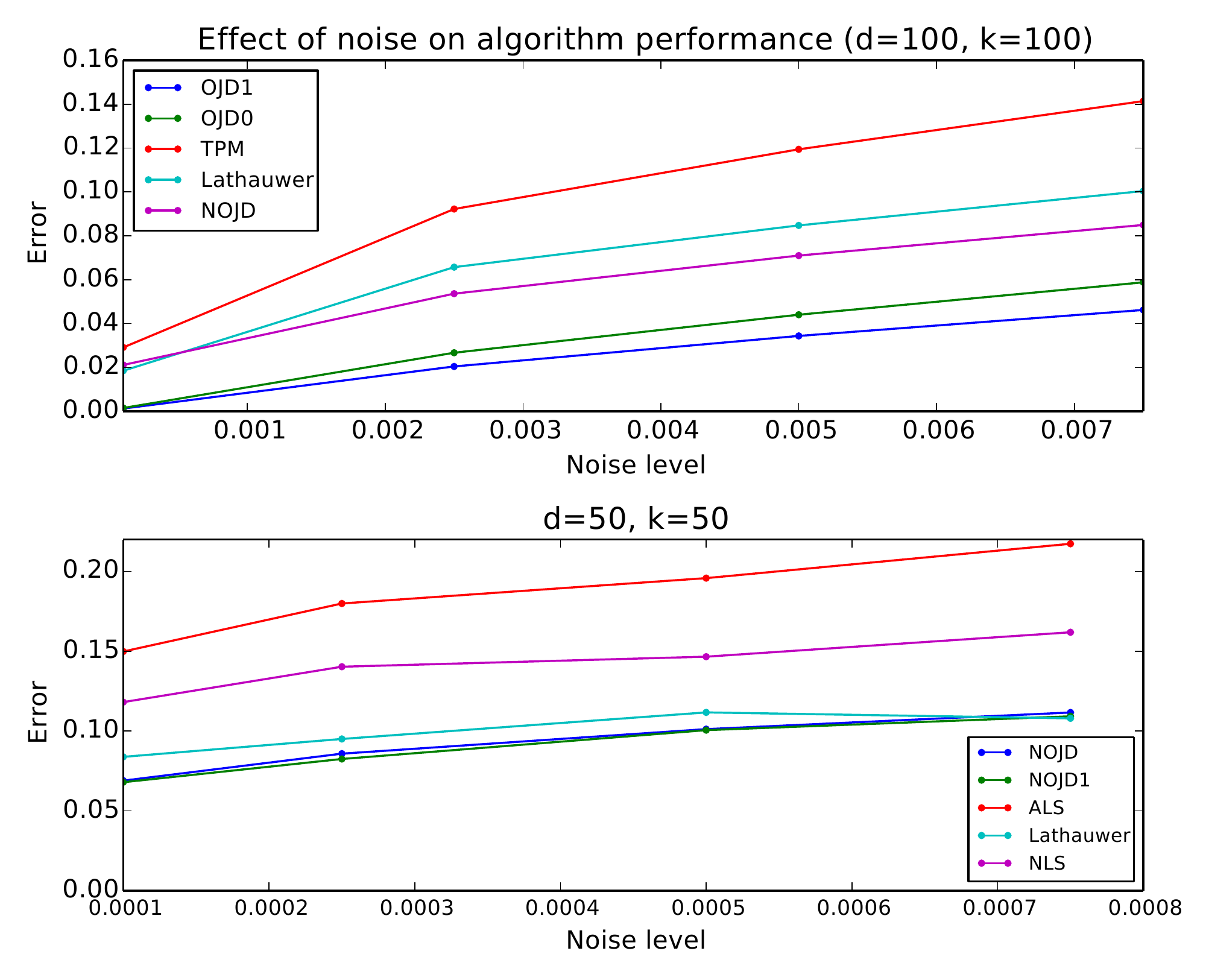}
\end{center}
\caption{Performance on full-rank synthetic tensors.}\label{fig:synth-comparison}
\end{figure}

%In the undercomplete case (Figure \ref{fig:snynth-comparison},top), all
%algorithms fare similarly and errors are within 10\% of each other.
Our method realizes its full potential in the full-rank non-orthogonal setting, where {\bf OJD0} and
{\bf OJD1} are up to three times more accurate than alternative methods (Figure
\ref{fig:synth-comparison}, top). In the (arguably easier) undercomplete case,
our methods do not achieve more than a 10\% improvement, and overall, all
algorithms fare similarly (Figure \ref{fig:ortho-comparison} in the supplementary material). Alternating least squares displayed very poor performance, and we omit it from our graphs.

%\begin{figure}
%\begin{center}
%\includegraphics[width=8cm]{figures/ortho-comparison-latest.pdf}
%\end{center}
%\caption{Algorithm performance on synthetic orthogonal tensors.}\label{fig:orthogonal-comparison}
%\end{figure}

%We follow the same experimental setup as above and summarize our experiments in Figure
%\ref{fig:nonortho-comparison}.
%In the undercomplete setting, Lathauwer's algorithm has the
%highest accuracy, about a 10\% more than our approach (Figure
%\ref{fig:nonortho-comparison}, right). 
In the full rank setting, there is
little difference in performance between our method and {\bf Lathauwer} (Figure \ref{fig:synth-comparison}, bottom).
In both the full and low-rank cases (Figure \ref{fig:synth-comparison}, bottom and Figure \ref{fig:nonortho-comparison} in the supplementary material), we consistently outperform the standard approaches, {\bf ALS} and {\bf NLS}, by 20--50\%.
Although we do not always outperform {\bf Lathauwer} (a state-of-the-art method), {\bf NOJD0} and {\bf NOJD1}
are faster and much simpler to implement.

%\begin{figure}
%\begin{center}
%\includegraphics[width=8cm]{figures/nonortho-comparison-latest.pdf}
%\end{center}
%\caption{Algorithm performance on synthetic non-orthogonal tensors.}\label{fig:nonorthogonal-comparison}
%\end{figure}

% Method of moments
We also tested our method on the single topic model from Section \ref{sec:mom}.
For $d=50$ and $k=10$, over 50 trials in which model parameters were generated uniformly at random in $\mathcal S^{d-1}$,
{\bf OJD0} and {\bf OJD1}
obtained error rates of $0.05$ and $0.055$ respectively,
followed by {\bf TPM} ($0.62$ error),
and {\bf Lathauwer} ($0.65$ error).
Additional experiments
on asymmetric tensors and on running time are in the supplementary material.

\paragraph{Community detection in a social network}

\begin{figure}
\begin{center}
%\framebox[4.0in]{$\;$}
\includegraphics[width=8cm]{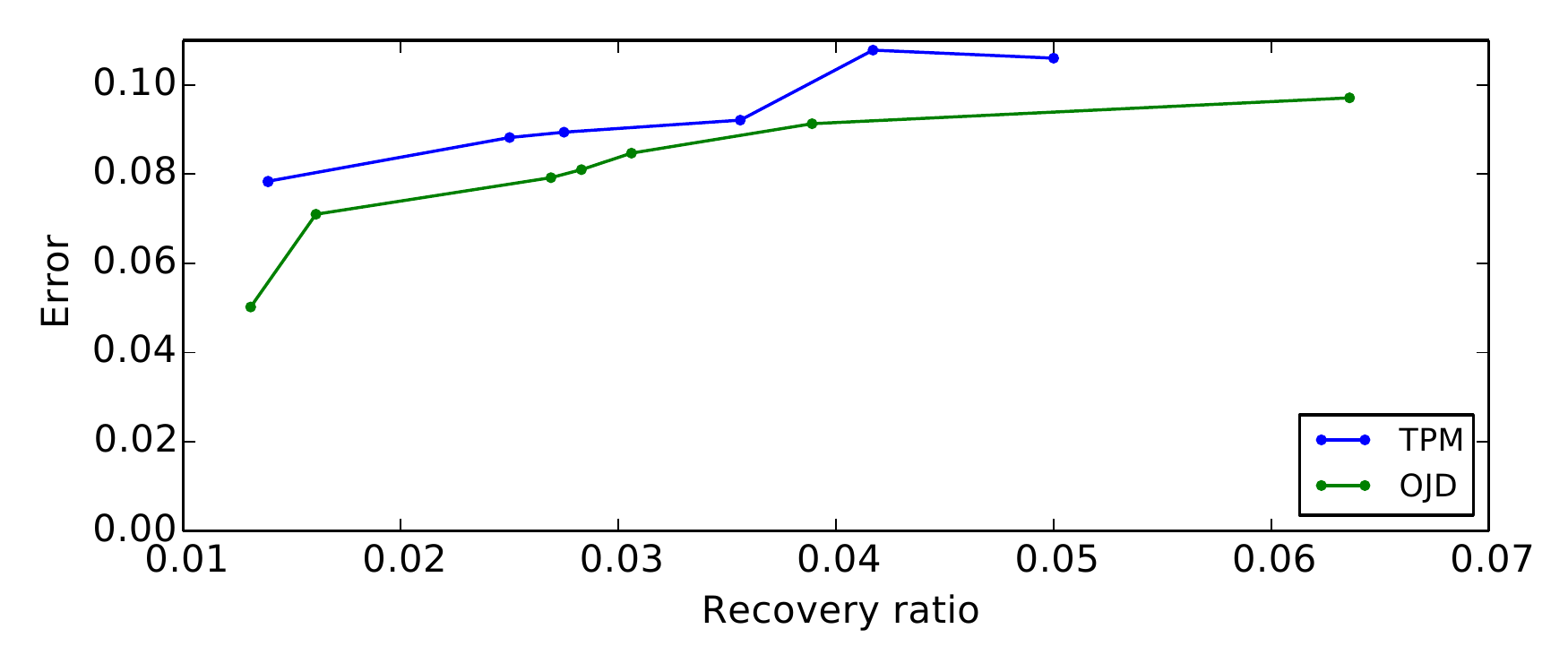}
\end{center}
\caption{Accuracy/recovery tradeoff for community detection.}\label{fig:fb-curve}
\end{figure}

Next, we use our method to detect communities in a real Facebook friend network
at an American university \cite{huang2013fast} using a recently developed
estimator based on the method of moments \cite{anandkumar2013community}. We
reproduce a previously proposed methodology for assessing the performance of
this estimator on our Facebook dataset \cite{huang2013fast}: ground truth
communities are defined by the known dorm, major, and high school of each
student; empirical and true community membership vectors $\hat c_i, c_i$ are
matched using a similarity threshold $t > 0$; for a given threshold, we define
the {\em recovery ratio} as the number of true $c_i$ to which an empirical
$\hat c_i$ is matched and we define the {\em accuracy} to be the average
$\ell_1$ norm distance between $c_i$ and all the $\hat c_i$ that match to it.
See \cite{huang2013fast} for more details. By varying $t>0$, we obtain a
tradeoff curve between the recovery ratio and accuracy (Figure
\ref{fig:fb-curve}). Our {\bf OJD1} method determines the top $10$ communities more accurately than {\bf TPM};
finding smaller communities was
equally challenging for both methods.
%We did not test non-orthogonal methods on
%this task, as the unwhitened tensor is too large to be decomposed by such methods.

\paragraph{Label prediction from crowdsourcing data}

Lastly, we predict data labels within
several datasets based on real-world crowdsourcing annotations  
using  a recently proposed estimator based on the
method of moments \cite{zhang2014crowdsourcing}. 
We incorporate our tensor factorization algorithms
within the estimator and
evaluate the
approach on the same datasets as \cite{zhang2014crowdsourcing} except
one, which we could not obtain. In addition to the previously defined methods, we also compare to the
expectation maximization algorithm initialized with majority voting by the
workers ({\bf MV+EM}). We measure the label prediction accuracy.
Overall, {\bf NOJD1} outperforms all other tensor-based methods on three out of four
datasets and results in accuracy gains of up to $1.75\%$ (Table
\ref{tab:crowdsourcing}). {\bf OJD1} outperforms the {\bf TPM} on every
dataset but one, and in two cases even outperforms {\bf ALS} and {\bf Lathauwer}, even though
they are not affected by whitening. Most interestingly, on two datasets, at
least one of our methods matches or outperforms the EM-based estimator.
% PL: this is pretty weak / grim for tensor methods
%This suggests that our algorithm may improve a method
%of moments estimator to the point of making it competitive in EM on a
%real-world task.

\begin{table}
\caption{Crowdsourcing experiment results}\label{tab:crowdsourcing}
\begin{center}
\begin{tabular}{c|c c c c c}
Dataset		& Web	& RTE		& Birds		& Dogs \\
\hline
TPM		& 82.25		& 88.75		& 87.96		& 84.01 \\
OJD		& 82.33		& 90.00		& {\bf 89.81}	& 84.01 \\
NOJD	& {\bf 83.49}	& {\bf 90.50}	& {\bf 89.81}	& {\bf 84.26} \\
ALS		& 83.15		& 88.75		& 88.89		& {\bf 84.26} \\
LATH	& 83.00		& 88.75		& 88.89		& {\bf 84.26} \\
\hline
MV+EM	& {\bf 83.68}	& {\bf 92.75}	& 88.89	& 83.89 \\
\hline					
 Size &	2665 & 	800	& 106	& 807
\end{tabular}
\end{center}
\end{table}

\section{Discussion}
\label{sec:conclusion}

% High-level: matrix algorithms > tensor algorithms. 

% Tensor factorization 
% HOSVD is one of the most popular tensor factorization algorithms with
% many uses - it is simple: combine the SVDs of a number of matrix
% unfoldings of a tensor. Likewise, we propose an algorithm that solves
% the ``eigendecomposition'' problem for tensors by reducing it to
% simultaneous matrix decomposition.\reword

We have presented a simple method for tensor
  factorization based on three ideas:
simultaneous matrix diagonalization, random projections, and plugin estimates.
% Contrast with what exists:
%   - simultaneous diagonalization is heavily used; previous reductions required solving a huge system, which we avoid through projection.
Joint diagonalization methods for tensor
  factorization have been proposed in the past, but they have either been
  computationally too expensive \cite{delathauwer2006decomposition} or
  numerically unstable \cite{anandkumar12moments}.
We overcome both these limitations using multiple random projections
  of the tensor.
%   - random projection 
Note that our use of random projections is atypical: instead of using 
  projections for dimensionality reduction (e.g.
  \cite{halko2011structure}), we use it to reduce the {\em order} of the tensor.
% --
Finally, we improve estimates of the factors retrieved
  with random projections by using them as plugin estimates, a common
  technique in statistics to improve statistical efficiency
\cite{vaart98asymptotic}. %\reword
Extensive experiments show that our
factorization algorithm is more accurate than the
state-of-the-art.

\bibliographystyle{unsrtnat}
\bibliography{all}

\newpage
\appendix
\onecolumn

% AC: Made redundant.
%\section{Simultaneous diagonalization}
%\label{sec:simdiagDetails}
%
%\pl{High-level: what's the point of this?
%In general, try not to make appendices a dumping ground;
%they should be clearly written and easy to read
%}
%
%Recall that $V^{-1} = B_T \cdots B_2 B_1$,
%where at each iteration $t = 1, \dots, T$, we choose $B_t$ to minimize $J(V)$.
%
%The Jacobi algorithm for the orthogonal case is a simple adaptation of the
%Jacobi method for diagonalizing a single matrix.
%Here, each $B_t$ is chosen to be a rotation defined by two of the $d$ axes $i < j \in [d]$:
%$B_t = I + (\cos\theta - 1) (\delta_{ii} + \delta_{jj}) + \sin\theta
%(\delta_{ij} - \delta_{ji})$ for some angle $\theta$ \pl{use $\Delta$ to have consistent notation}.
%We sweep over all $i<j$,
%compute the best angle $\theta$ in closed form using \citet{cardoso1996joint}
%to obtain $B_t$, and update each $M_l$ by $B_t M_l B_t^\top$.
%The above can be done in $O(d^3 L)$ time per sweep.
%
%\pl{does this actually convey more information than the main paper?
%doesn't seem like it's the case}
%
%For the non-orthogonal case, the LUJ1D algorithm is similar,
%  expect that $B_t$ is chosen to be either a lower or upper unit triangular matrix
%  ($B_t = I + a \delta_{ij}$ for some $a$ and $i \neq j$).
%The optimal value of $a$ that minimizes $J(V)$ can also be computed in closed form
%(see \cite{afsari2006simple} for details).
%The running time per iteration is the same as before ($O(d^3 L)$).

\section{Experiments}

\subsection{Synthetic experiments}

\paragraph{Orthogonal tensors}

\begin{figure}
\begin{center}
%\framebox[4.0in]{$\;$}
\includegraphics[width=8cm]{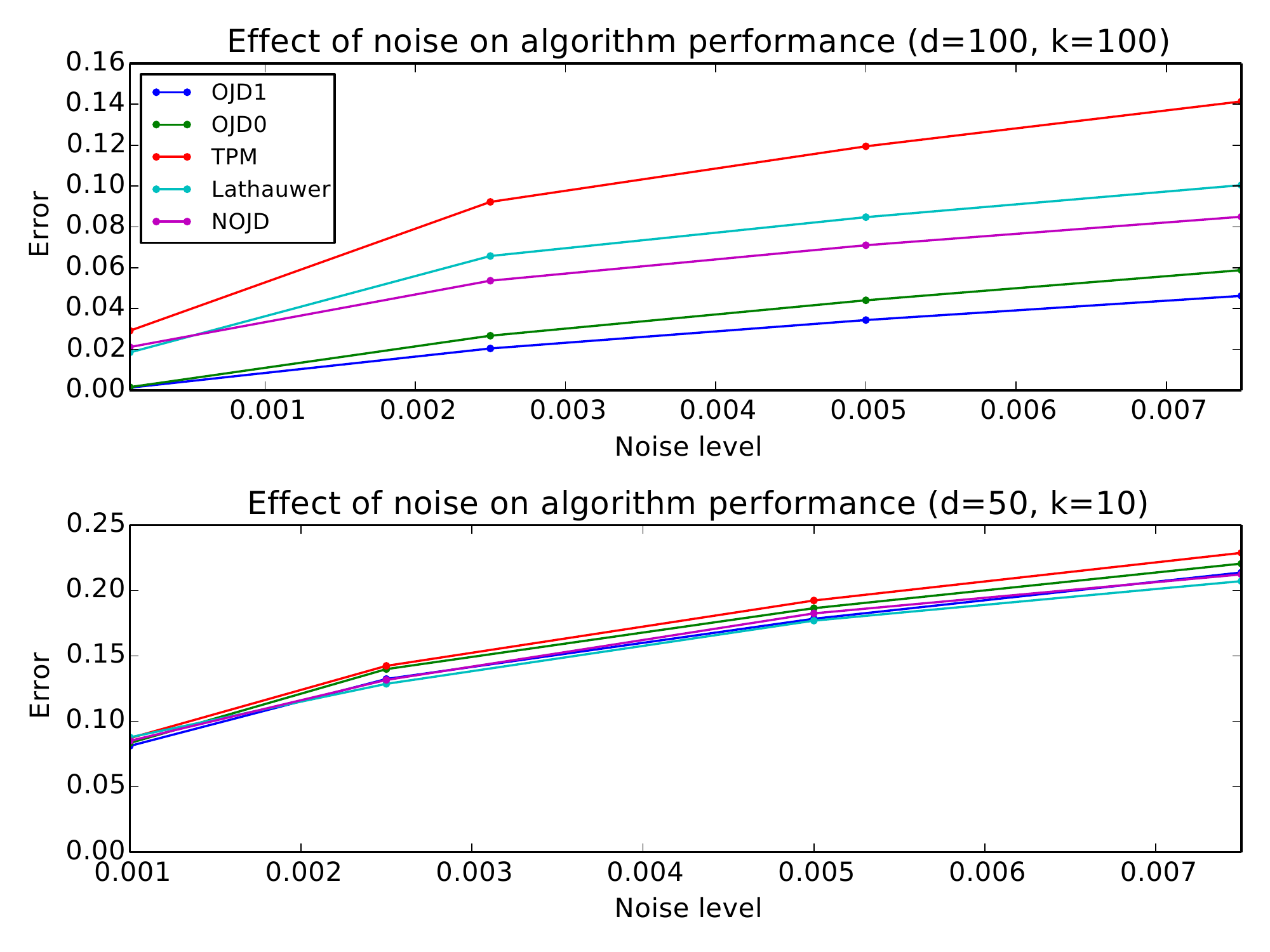}
\end{center}
\caption{Algorithm performance in the orthogonal setting.}\label{fig:ortho-comparison}
\end{figure}

We start by generating random tensors $T = \sum_i \pi u_i^{\otimes 3} + \e R$ with Gaussian entries in
$\pi, R$ and $u_i$
distributed uniformly in the unit sphere $\mathcal S^{d-1}$.
We let $d=25, 50, 100$ and in each case consider two
regimes: undercomplete tensors with $k=0.2d$ and full rank tensors, $k=d$. 
We vary $\e$ and report the average error $\|\ut_i -
u_i\|_2$ across all eigenvectors $u_i$ and across 50 trials.
In the orthogonal setting,
we compare our algorithms (\textbf{OJD0} uses random projections, \textbf{OJD1} is with plugin)
with the tensor power method (\textbf{TPM}), alternating least squares (\textbf{ALS}), and with
the method of de \textbf{Lauthauwer} \cite{delathauwer2006decomposition}.
Alternating least squares displayed very poor performance, and we omit it from our graphs.
In the undercomplete case (Figure \ref{fig:ortho-comparison}, right), all
algorithms fare similarly and errors are within 10\% of each other.
Our method realizes its full potential in the full-rank setting, where OJD0 and OJD1 are up to three times more accurate than alternative methods ((Figure \ref{fig:ortho-comparison}, left).

\paragraph{Non-orthogonal tensors}

\begin{figure}
\begin{center}
%\framebox[4.0in]{$\;$}
\includegraphics[width=8cm]{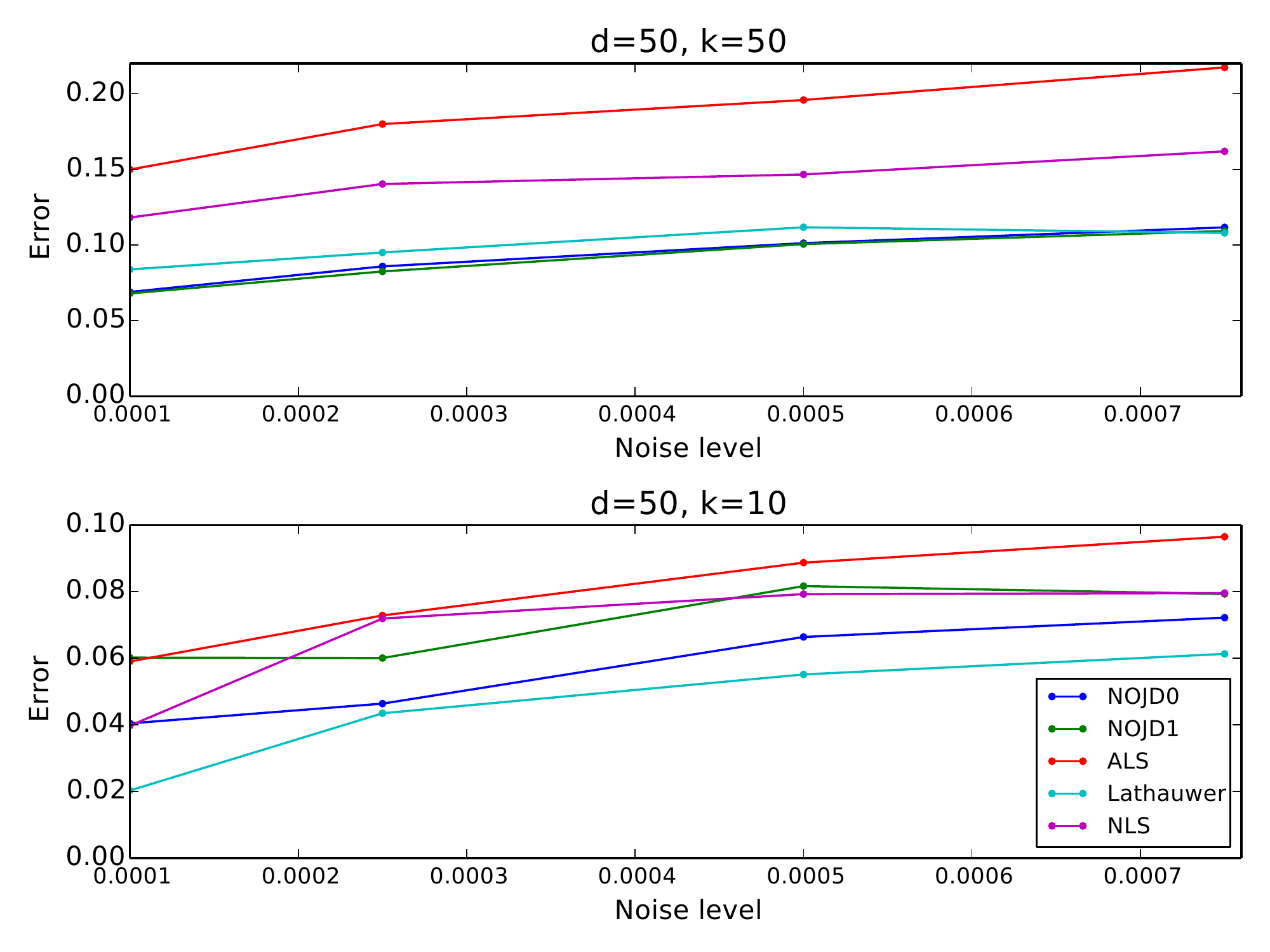}
\end{center}
\caption{Algorithm performance in the non-orthogonal setting.}\label{fig:nonortho-comparison}
\end{figure}

In the non-orthogonal setting, we compare de \textbf{Lathauwer},
alternating least squares (\textbf{ALS}), non-linear least squares (\textbf{NLS}),
and our non-orthogonal methods (\textbf{NOJD0} and \textbf{NOJD1}).
We follow the same experimental setup as above and summarize our experiments in Figure
\ref{fig:nonortho-comparison}.
In the undercomplete setting, Lathauwer's algorithm has the
highest accuracy, about a 10\% more than our approach (Figure
\ref{fig:nonortho-comparison}, right). 
In the full rank setting, there is
little difference in performance between our method and Lathauwer's.
In both settings, we consistently outperform the standard approaches, ALS and NLS, by 20-50\% (Figure \ref{fig:nonortho-comparison}, left).
Although we do not always outperform Lauthauwer's state-of-the-art method, NOJD0 and NOJD1
are faster and much simpler to implement.

\paragraph{Asymmetric tensors}

\begin{figure}
\begin{center}
%\framebox[4.0in]{$\;$}
\includegraphics[width=8cm]{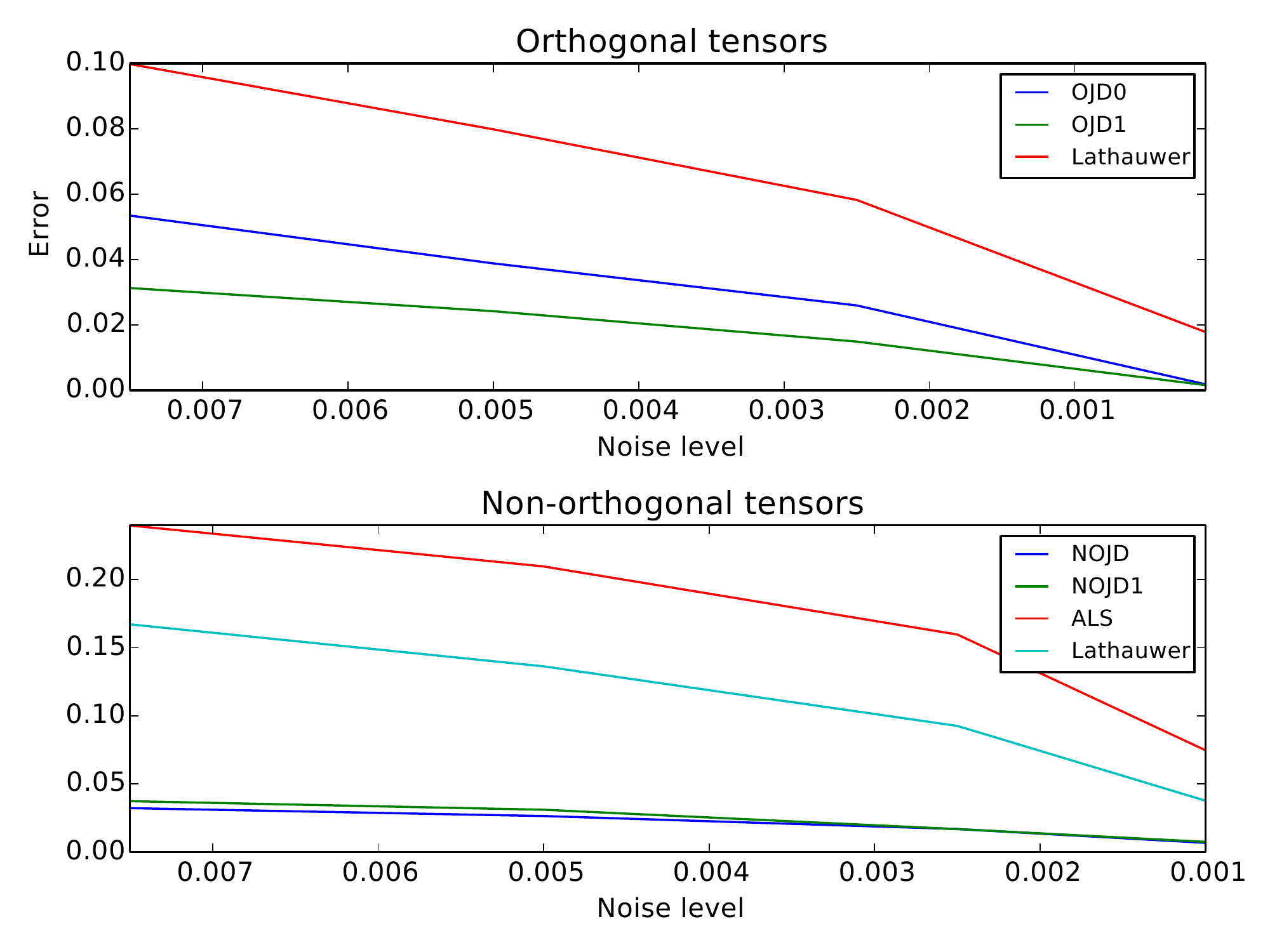}
\end{center}
\caption{Algorithm performance on asymmetric tensors.}\label{fig:svd-comparison}
\end{figure}

Lastly, we evaluate the extension of our algorithm to tensors of size $50
\times 50 \times 50$ having three distinct sets of asymmetric components (one
in each mode). We find that performance is  consistent with the symmetric
setting, in both orthogonal and non-orthogonal regimes; our method outperforms
is competitors by at least $25\%$, and in the non-orthogonal setting, it
achieves an error reduction of up to 70\% over Lathauer (Figure
\ref{fig:svd-comparison}).

\subsection{Algorithm running time}\label{sec:runtime}

\begin{figure}
\begin{center}
%\framebox[4.0in]{$\;$}
\includegraphics[width=8cm]{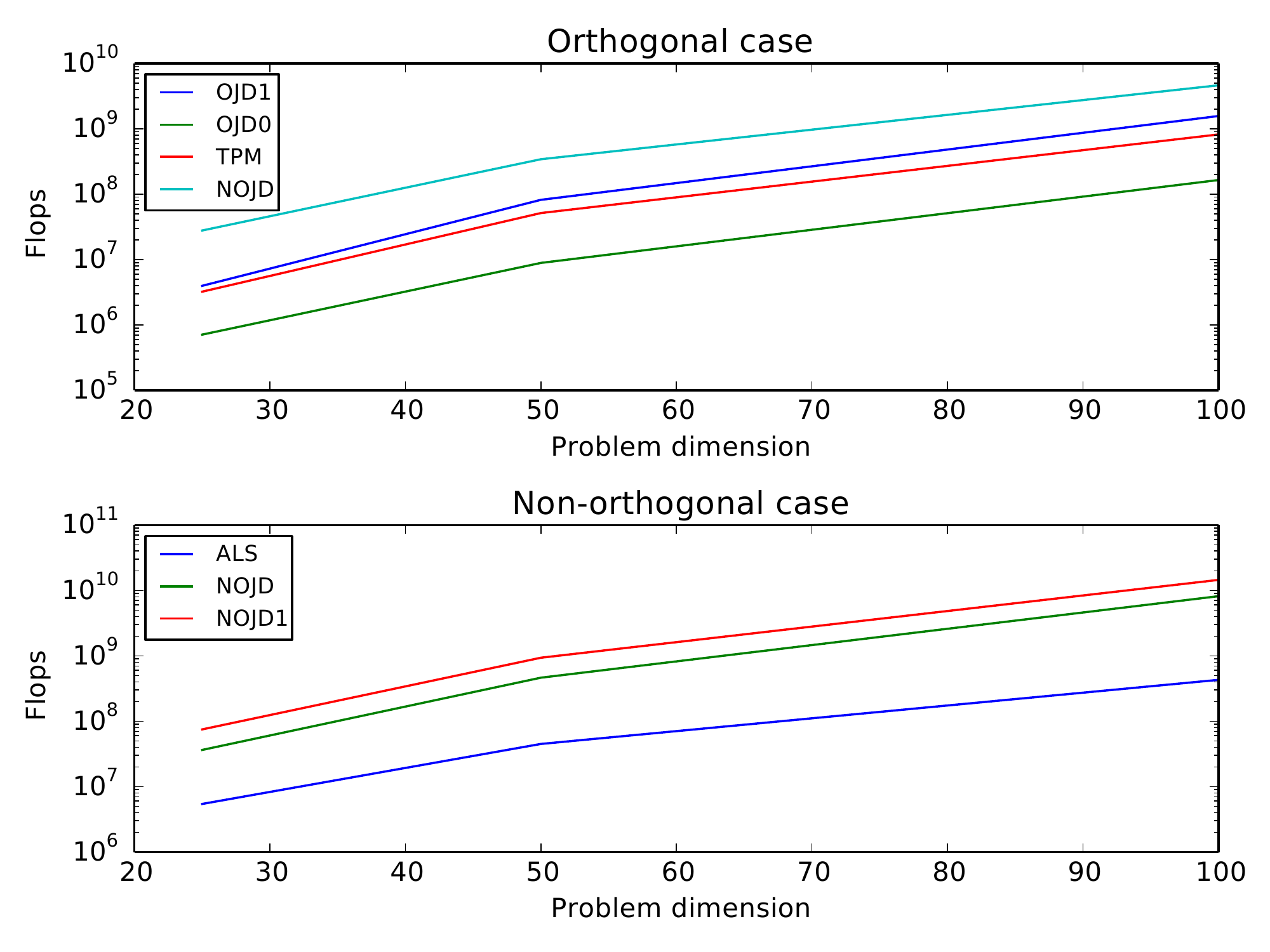}
\end{center}
\caption{Number of flops performed by various algorithms.}\label{fig:time-comparison}
\end{figure}

Figure \ref{fig:time-comparison} compares the running time in flops of the main
algorithms.

We obtain the plots in Figure \ref{fig:time-comparison} by calculating flops as
follows. The Jacobi method performs at each sweep $2dL(dk - {k \choose 2})$
flops (where $L$ is the number of matrices); the QRJ1 non-orthogonal
diagonalization algorithm performs $4d^3L$ flops per sweep. The tensor power
method performs a total of $Lkd^3$ flops (where $L$ is the number of restarts),
times the number of steps it takes to reach convergence for a given
eigenvector. The flop count of Lathauwer's method is much higher than that of other method's: at one stage, it requires
finding the SVD of a $d^4 \times k^2$ matrix. Consequently, we do not include it in our summary.

\section{Proofs for orthogonal tensor factorization}
\label{app:ortho-proofs}

In this section we prove perturbation bounds for our algorithm in the
  setting of orthogonal tensors.

Recall that we observe $\Th = T + \e R$ where $T = \sum_{i=1}^k \pi_i
u_i\tp{3}$ where $\pi_i$ are factor weights, $u_i \in \Re^d$ are orthogonal unit vectors and $R$ is, without
loss of generality, symmetric with $\|R\|_\op = 1$.
Our objective is to estimate $\pi$ and $(u_i)$. 
\algorithmref{joint} does so by simultaneously diagonalizing a number of
  projections of $T$;
we make use of projections along random vectors and along approximate
factors. In this section we will show why both schemes recover
$\pi_i$ and $(u_i)$ with high probability.

\paragraph{Setup}
Let $\sM = \{M_1, \ldots, M_L\}$ be the projections of $T$ along vectors
  $w_1, \ldots, w_L$, and $\sMh = \{ \Mh_1, \ldots, \Mh_L \}$ be the projections
  of $\Th$ along $w_1, \ldots, w_L$.
We have that $M_l = \sum_{i=1}^d \pi_i (w_l^\top u_i) u_i \otimes u_i$
  and that $\Mh_l = M_l + \epsilon R_l$, where $R_l = R(I,I,w_l)$.
Thus, $M_l$ are a set of simultaneously diagonalizable matrices with factors $U$ and factor weights $\lambda_{il} \eqdef \pi_i (w_l^\top u_i)$. 
From the discussion in \sectionref{background}, let $\Ub$ be a full-rank extension of $U$, with columns $u_1, u_2, \dots u_d$. 

  Let $\pit$ and $\ut$ be a factorization of $\Th$ returned by \algorithmref{joint}.
  From \lemmaref{cardoso}, we have that 
\begin{align}
  \| \ut_j - u_j \|_2 &\leq \e \sqrt{\sum_{i=1}^d E_{ij}^2} + o(\e), \label{eqn:ortho-perturb-1}
\end{align}
for $j \in [k]$ where $E \in \mathbb R^{d \times k}$ has entries
\begin{align}
  E_{ij} 
  &= 
  \begin{cases}
    0 & \text{for}~ i = j \\
    \frac{\sum_{l=1}^L (\lambda_{il} - \lambda_{jl}) u_j^\top R_l u_i}
    {\sum_{l=1}^L (\lambda_{il} - \lambda_{jl})^2} & \text{for}~ i \neq j.
  \end{cases}
\end{align}
%and $\lambda_{il}$ is the $i$-th eigenvalue of matrix $M_l$.

%Recall that $\lambda_{il}$ is the $i$-th eigenvalue corresponding to the $l$-th projection of the original tensor, 
%so $\lambda_{il} = \pi_i w_l^\top u_i$.
% PL: not necessarily random
For notational convenience, let $p_{ij} \eqdef (\pi_{i} u_i - \pi_{j} u_j)$ so that $\lambda_{il} - \lambda_{jl} = w_l^\top p_{ij}$.
Let $r_{ij} \eqdef R(u_i, u_j, I)$ so that 
\begin{align*}
  u_j^\top R_{l} u_i
    &= R(u_j, u_i, w_l) 
    = R(u_i, u_j, I)^\top w_l 
    = r_{ij}^\top w_l.
\end{align*}

The expression for $E_{ij}$ when $j \neq i$ simplifies to,
\begin{align}
  E_{ij} 
  &= \frac{\sum_{l=1}^L w_l^\top p_{ij} r_{ij}^\top w_l}
      {\sum_{l=1}^L w_l^\top p_{ij} p_{ij}^\top w_l}. \label{eqn:e-ij-form}
\end{align}

In the rest of this section, we will bound $E_{ij}$ for different choices of $\{w_l\}_{l=1}^L$.
%\pl{this is nice and clear}

\subsection{Plugin projections}

In \sectionref{orthogonal} we proposed using approximate factors $\ut_i$
as directions to project the tensor $\Th$ along.
In this section, we show that doing so guarantees small errors in $u_i$.

We begin by bounding the terms $E_{ij}$. 
\begin{lemma}[$E_{ij}$ with plug-in projections]
  \label{lem:e-ij-plugin}
  Let $w_1, \ldots, w_k$ be unit-vectors approximations of the unit vectors $u_1, \ldots, u_k$: $\|w_l - u_l\|_2 \leq \gamma$ (so $L = k$),
  and let $\sMh = \{\Mh_1, \ldots,
  \Mh_L\}$ be constructed via projection of $\Th$ along $w_1, \ldots,
  w_L$.
%  \pl{recall $u_j$ are unit vectors}
If the set of matrices $\sMh$ is simultaneously diagonalized, then to a first-order approximation,
\begin{align*}
  E_{ij} &= 
  \frac{ p_{ij}^\top r_{ij}}{\|p_{ij}\|^2} 
  + O(\gamma).
\end{align*}
\end{lemma}
\begin{proof}
  We have that 
  \begin{align*}
  w_l^\top (p_{ij}) 
    &= (u_l + (w_l - u_l))^\top (\pi_i u_i - \pi_j u_j) \\
    &= \pi_i \delta_{il} - \pi_j \delta_{jl} + (w_l - u_l)^\top (\pi_i u_i - \pi_j u_j) \\
    &\le \pi_i \delta_{il} - \pi_j \delta_{jl} + \|w_l - u_l\|_2 \|\pi_i u_i - \pi_j u_j\|_2 \\
    %&= \pi_i \delta_{il} - \pi_j \delta_{jl} + \|w_l - u_l\|_2 \frac{ (w_l - u_l)^\top (\pi_i u_i - \pi_j u_j) } { \|w_l - u_l\|_2 } \\
   &= \pi_i \delta_{il} - \pi_j \delta_{jl} + O(\gamma),
  \end{align*}
  where $\delta_{ij} = 1$ if $i = j$ and $0$ otherwise.

  Thus, 
  \begin{align*}
    E_{ij} 
    &= \frac{\sum_{l=1}^L w_l^\top p_{ij} r_{ij}^\top w_l}
        {\sum_{l=1}^L w_l^\top p_{ij} p_{ij}^\top w_l} \\
        &= \frac{\sum_{l=1}^L \left(\pi_{i} \delta_{il} - \pi_j \delta_{jl} + O(\gamma) \right) r_{ij}^\top w_l }
        {\sum_{l=1}^L (\pi_{i} \delta_{il} - \pi_j \delta_{jl} + O(\gamma) )^2} \\
        &= \frac{ \pi_i r_{ij}^\top w_i - \pi_j r_{ij}^\top w_j  + O(\gamma) }
        {\pi_i^2 + \pi_j^2 + O(\gamma)} \\
        &= \frac{ \pi_i r_{ij}^\top u_i + \pi_i (w_i - u_i)^\top r_{ij} - \pi_j r_{ij}^\top u_j - \pi_j (w_j - u_j)^\top r_{ij} + O(\gamma) }{\pi_i^2 + \pi_j^2 + O(\gamma)} \\
  \end{align*}
  Note that $(w_i - u_i)^\top r_{ij} = O(\gamma)$ and $(w_j - u_j)^\top
  r_{ij} = O(\gamma)$, and hence both can be included in the $O(\gamma)$ term.
  \begin{align*}
    E_{ij} 
        &= \frac{ r_{ij}^\top (\pi_i u_i - \pi_j u_j) + O(\gamma) }{\pi_i^2 + \pi_j^2 + O(\gamma)}.
  \end{align*}
  Finally, recall that $p_{ij} \eqdef (\pi_i u_i - \pi_j u_j)$ and that
  $\|p_{ij}\|^2 = \pi_i^2 + \pi_j^2$. Combining this with the observation that $\frac{1}{1-x} = 1 + x + o(x)$, we obtain
  \begin{align*}
    E_{ij} 
        &= \frac{ p_{ij}^\top r_{ij} }{\|p_{ij}\|^2} + O(\gamma).
  \end{align*}
\end{proof}

Next, we use these term-wise bounds to bound the error in $u_i$.
\begin{theorem}[Tensor factorization with plugin projections]
  \label{thm:ortho-fact-plugin-1}
  Let $w_1, \ldots, w_k$ be approximations of $u_1, \ldots, u_k$ such that $\|w_l - u_l\|_2 \le \gamma = O(\e)$,
    and let $\sMh = \{\Mh_1, \ldots, \Mh_L\}$ be constructed via
    projection of $\Th$ along $w_1, \ldots, w_L$.
  Then, for $j \in [k]$,
  \begin{align*}
    \| \ut_j - u_j \|_2 
    &\leq 
    \left(\frac{2\sqrt{\|\pi\|_1 \pi_{\max}}}{\pi_{i}^2}\right) \e 
     + o(\e).
  \end{align*}
\end{theorem}
\begin{proof}
  From \equationref{ortho-perturb-1}, we have that,
  \begin{align*}
    \| \ut_j - u_j \|_2 &\leq \e \sqrt{\sum_{j=1; j \neq i}^d E_{ij}^2},
  \end{align*}
  for all $j \in [k]$.
  By \lemmaref{e-ij-plugin}, we get,
  \begin{align*}
    E_{ij} &= \frac{p_{ij}^\top r_{ij}}{\|p_{ij}\|^2} + O(\e),
  \end{align*}
  and thus,
  \begin{align*}
    \| \ut_j - u_j \|_2 
      &\leq \e \sqrt{
      \sum_{i = 1; i \neq j}^{d} \left( \frac{p_{ij}^\top r_{ij}}{\|p_{ij}\|^2} \right)^2 } + o(\e).
  \end{align*}

  Now, we must bound $\sum_{i=1;i\neq j}^d (p_{ij}^\top r_{ij})^2$. We
  expect this the projection to mostly preserve the norm of $p_{ij}$
  because $r_{ij}$ are effectively random vectors. 
  Using \lemmaref{bounding-pi} with $\mu = 0$, we get that
  $\sum_{i=1;i\neq j}^d (p_{ij}^\top r_{ij})^2 \le 4 \|\pi\|_1\pi_{\max}
  $. Finally, $\|p_{ij}\|_2^2 = \pi_i^2 + \pi_j^2 \ge \pi_{j}^2$.
  \begin{align*}
    \| \ut_j - u_j \|_2 
      &\leq  
      \left(\frac{\sqrt{4 \| \pi \|_1 \pi_{\max}}}{\pi_{j}^2 }\right) \e
            + o(\e) \\
      &\leq 
      \left(\frac{2\sqrt{\|\pi\|_1 \pi_{\max}}}{\pi_{j}^2}\right) \e  + o(\e).
  \end{align*}
\end{proof}

\subsection{Random projections}
Let us now consider the case when $\{w_l\}_{l=1}^L$ are random Gaussian
vectors and present similar bounds. 

Given \equationref{e-ij-form}, we should expect $E_{ij}$ to sharply, and
now show that this is indeed the case. 
\begin{lemma}[Concentration of error $E_{ij}$]
  \label{lem:e-ij-conc}
Let $w_1, \ldots, w_L$ be i.i.d.~random Gaussian vectors $w_l \sim
  \sN(0, I)$, and let $\sMh = \{\Mh_1, \ldots,
  \Mh_L\}$ be constructed via projection of $\Th$ along $w_1, \ldots,
  w_L$.
If the set of matrices $\sMh$ is simultaneously diagonalized, then the
  first-order error $E_{ij}$ is sharply concentrated. If $L \ge 16 \log(2\delta)$, then with probability at least $1 - \delta$, 
  \begin{align*}
    E_{ij} 
        &\le
        \frac{p_{ij}^\top r_{ij}}{\|p_{ij}\|_2^2} 
            + \frac{10 \log(2/\delta)}{\sqrt L} \frac{\|r_{ij} \|_2}{\|p_{ij}\|_2}.
      \end{align*}
\end{lemma}
\begin{proof}
  The numerator and denominator of \equationref{e-ij-form} are both
  distributed as the sum of $\chi^2$ variables; we show below that they respectively concentrate about
  $p_{ij}^\top r_{ij}$ and $\|p_{ij}\|^2_2$. 

  From \lemmaref{gaussian-prod}, we have that the following hold
  independently with probability at least $1-\delta/2$,
  \begin{align*}
    \frac{1}{L} \sum_{l=1}^{L} w_l^\top p_{ij} r_{ij}^\top w_l 
    &\le p_{ij}^\top r_{ij} + \|p_{ij}\| \|r_{ij}\| \left( 3 \sqrt{\frac{\log(2/\delta)}{L}} \right) \\
    \frac{1}{L} \sum_{l=1}^{L} w_l^\top p_{ij} p_{ij}^\top w_l 
    &\ge \|p_{ij}\|^2 \left( 1 - \frac{2\log(2/\delta)}{\sqrt{L}} \right) \\
  \end{align*}
  Applying a union bound on both these events, we get that with probability at least $1 - \delta$,
  \begin{align*}
    E_{ij} &=
    \frac{\sum_{l=1}^L w_l^\top p_{ij} r_{ij}^\top w_l}
            {\sum_{m=1}^L \|w_m^\top p_{ij}\|_2^2}  \\
            &\le 
            \frac
            {p_{ij}^\top r_{ij} + \|p_{ij}\|_2 \|r_{ij}\|_2 \left( 3\sqrt{\frac{\log(2/\delta)}{L}} \right) }
            {\|p_{ij}\|_2^2 \left(1 - \frac{2 \log(2/\delta)}{\sqrt{L}} \right)}.
  \end{align*}

  Note that with the given condition on $L$, $\frac{2
  \log(2/\delta)}{\sqrt{L}} < \half$. Using the property that when $x
  \le \half$, $\frac{1}{1 - x} \le 1 + 2x$, 
  we have that 
  \begin{align*}
    \frac{1}{1 - \frac{2 \log(2/\delta)}{\sqrt{L}}}
      &\le 1 + \frac{4 \log(2/\delta)}{\sqrt{L}}.
  \end{align*}

  Consequently,
  \begin{align*}
    E_{ij} &\le 
    \frac{1}{\|p_{ij}\|_2^2} 
      \left({p_{ij}^\top r_{ij} + \|p_{ij}\|_2 \|r_{ij}\|_2 \left( 3\sqrt{\frac{ \log(2/\delta)}{L}} \right) } \right)
          \left(
          {1 + \frac{4 \log(2/\delta)}{\sqrt{L}}}
          \right) \\
        &\le
        \frac{p_{ij}^\top r_{ij}}{\|p_{ij}\|_2^2} \left(1 + \frac{4\log(2/\delta)}{\sqrt{L}}\right)
            + 6 \frac{\|r_{ij} \|_2}{\|p_{ij}\|_2} \sqrt{\frac{\log(2/\delta)}{L}} \\
        &\le
        \frac{p_{ij}^\top r_{ij}}{\|p_{ij}\|_2^2} 
            + \frac{10 \log(2/\delta)}{\sqrt L} \frac{\|r_{ij} \|_2}{\|p_{ij}\|_2}.
  \end{align*}
\end{proof}

With this term-wise bound, we can again proceed to bounding the error $u_i$.
\begin{theorem}[Tensor factorization with random projections]
  \label{thm:ortho-fact-random-1}
  Let $w_1, \ldots, w_L$ be i.i.d.~random Gaussian vectors, $w_l \sim
    \sN(0, I)$, and let $\sMh = \{\Mh_1, \ldots,
    \Mh_L\}$ be constructed via projection of $\Th$ along $w_1, \ldots,
    w_L$.
  Furthermore, let $L \ge 16 \log(2 d (k-1)/\delta)^2$, then, with probability at least $1-\delta$,
  \begin{align*}
    \| \ut_j - u_j \|_2 
      &\leq 
      \left(\frac{2\sqrt{2\|\pi\|_1 \pi_{\max}}}{\pi_{i}^2}\right) \e 
      + \left(20 \sqrt{2} \log(2d(k-1)/\delta) \frac{\sqrt{d/L}}{\pi_{i}} \right) \e 
        + o(\e).
  \end{align*}
  for all $j \in [k]$.
\end{theorem}
\begin{proof}
  From \equationref{ortho-perturb-1}, we have that,
  \begin{align*}
    \| \ut_j - u_j \|_2 &\leq \e \sqrt{\sum_{i=1;i \neq j}^d E_{ij}^2} + o(\e).
  \end{align*}

  By \lemmaref{e-ij-conc}, with probability at least $1 - \delta/(d(k-1))$, 
  \begin{align*}
    E_{ij} 
        &\le
        \frac{|p_{ij}^\top r_{ij}|}{\|p_{ij}\|_2^2} 
            + \frac{10 \log(2d(k-1)/\delta)}{\sqrt L} \frac{\|r_{ij} \|_2}{\|p_{ij}\|_2}.
  \end{align*}

  Applying a union bound over $(E_{ij})_{j\neq i}^d$, we have that with
  probability at least $1-\delta$,
  \begin{align*}
    \| \ut_j - u_j \|_2 
      &\leq \e \sqrt{\sum_{i=1;i\neq j}^d 2 \left( \frac{p_{ij}^\top r_{ij}}{\|p_{ij}\|_2^2} \right)^2} 
        + \e \frac{10\log(2d(k-1)/\delta)}{\sqrt{L}}
        \sqrt{\sum_{i=1;i\neq j}^d 2 \left(\frac{\|r_{ij} \|_2}{\|p_{ij}\|_2} \right)^2}
        + o(\e),
  \end{align*}
  for all $j \in [k]$.
  We have used the fact that for $a, b \geq 0$, $(a+b)^2 = a^2 + 2ab
  + b^2 \leq a^2 + (a^2 + b^2) + b^2 = 2a^2 + 2b^2$ and $\sqrt{a
  + b} \le \sqrt{a} + \sqrt{b}$.
  
  Note that $\|p_{ij}\|_2 = \sqrt{\pi_{i}^2 + \pi_{j}^2} \ge |\pi_i|$.
  In \lemmaref{bounding-pi}, we show that $\sum_{i=1;i\neq j}^d (p_{ij}^\top
  r_{ij})^2 \le 4\|\pi\|_1 \pi_{\max}$. Furthermore, $\|r_{ij}\| \le 1$ by the
  operator norm bound on $R$. Thus, we get,
  \begin{align*}
    \| \ut_j - u_j \|_2 
      &\leq 
      \left(\frac{2\sqrt{2\|\pi\|_1 \pi_{\max}}}{\pi_{i}^2}\right) \e 
      + \left(20 \sqrt{2} \log(2d(k-1)/\delta) \frac{\sqrt{d/L}}{\pi_{i}} \right) \e 
        + o(\e).
  \end{align*}
\end{proof}

\section{Proofs for non-orthogonal tensor factorization}
\label{app:non-ortho-proofs}

In this section we extend our previous analysis to non-orthogonal tensor
decomposition.

\paragraph{Setup}
As before,
let $\sM = \{M_1, \ldots, M_L\}$ be the projections of $T$ along vectors
  $w_1, \ldots, w_L$, and $\sMh = \{ \Mh_1, \ldots, \Mh_L \}$ be the projections
  of $\Th$ along $w_1, \ldots, w_L$.
We have that $M_l = \sum_{i=1}^d \pi_i (w_l^\top u_i) u_i \otimes u_i$
  and that $\Mh_l = M_l + \epsilon R_l$, where $R_l = R(I,I,w_l)$.
Thus, $M_l$ are a set of simultaneously diagonalizable matrices with factors $U$ and factor weights $\lambda_{il} \eqdef \pi_i (w_l^\top u_i)$. Let $\Ub$ be the full-rank extension of $U$ with unit-norm columns $u_1, u_2, \dots, u_d$.
In this setting, however, the factor $U$ is not orthogonal. Let $\Vb = \Ub\inv$, with rows $v_1, v_2, \dots, v_d$. Note that we place our incoherence assumption on the columns of $U$ and present results in terms of the 2-norm of $V^\top$. When $U$ is incoherent, it can be shown that $\|V^\top\|_2 \le 1 + O(\mu)$.
Finally, note that in the orthogonal case, when $\mu = 0$, the rows $(v_i)$ and columns $(u_i)$ are
identical, and no distinction between the two need be made.

%\pl{does $u_i$ have to be unit norm?  why/when is the solution even unique?}

  Let $\pit$ and $\ut$ be a factorization of $\Th$ returned by \algorithmref{joint}.
From \lemmaref{afsari}, we have that 
$$ \|\ut_j - u_j\|_2 = \epsilon \sqrt{\sum_{i=1}^{d} E_{ij}^2},$$
where the entries of $E \in \mathbb R^{d \times k}$ are bounded by \lemmaref{no-eij-bound-1}:
\begin{align}
  |E_{ij}|
  &\le \frac{1}{1 - \rho_{ij}^2} 
  \left(\frac{1}{\|\lambda_i\|^2_2} + \frac{1}{\|\lambda_j\|^2_2}\right)
  \left( \left|\sum_{l=1}^L v_i^\top R_l v_j \lambda_{jl} \right| + \left|\sum_{l=1}^L v_i^\top R_l v_j \lambda_{il} \right| \right), \label{eqn:e-ij-bound}
\end{align}
%where $u_i$ are the {\em rows} of $U$,
where $\lambda_i \in \Re^L$ is the vector of $i$-th factor values of $M_l$, i.e. $\lambda_{il}$ is the $i$-th factor value of matrix $M_l$ (i.e. $\lambda_{il} = (\Lambda_l)_{ii}$) and
$\rho_{ij} = \frac{\lambda_{i}^\top \lambda_{j}}{\|\lambda_{i}\|_2
\|\lambda_{j}\|_2}$, the modulus of uniqueness, is a measure of
the singularity of the problem.

When $\lambda_{il}$ is generated by projections,
$\lambda_{il} = \pi_i w_l^\top u_i$.
Let $r_{ij} \eqdef R(v_i, v_j, I)$ so that 
\begin{align*}
  v_i^\top R_{l} v_j
    &= R(v_i, v_j, w_l) 
    = R(v_i, v_j, I)^\top w_l 
    = r_{ij}^\top w_l.
\end{align*}
Note that $\|r_{ij}\|_2 \le \|v_i\|_2 \|v_j\|_2 \le \|V^\top\|_2^2$.

\equationref{e-ij-bound} then simplifies to,
\begin{align}
  |E_{ij}| &\le \frac{1}{1 - \rho_{ij}^2} 
  \left(\frac{1}{\|\lambda_i\|^2_2} + \frac{1}{\|\lambda_j\|^2_2}\right)
  \left( |\pi_{j}| \left|\sum_{l=1}^L w_l^\top u_j r_{ij}^\top w_l \right| +
    |\pi_{i}| \left|\sum_{l=1}^L w_l^\top u_i r_{ij}^\top w_l \right|
    \right), \label{eqn:e-ij-non-ortho}
\end{align}
where $\|\lambda_i\|_2^2 = \pi_i^2 \sum_{l=1}^L w_l^\top u_i u_i^\top w_l$, and $\rho_{ij}$ has the following expression,
\begin{align}
  \rho_{ij} 
    &= \frac{\lambda_i^\top \lambda_j}{\|\lambda_i\|_2 \|\lambda_j\|_2 }
    = \frac{\sum_{l=1}^L w_l^\top u_i u_j^\top w_l}{
    \sqrt{ (\sum_{l=1}^L w_l^\top u_i u_i^\top w_l) (\sum_{l=1}^L w_l^\top u_j u_j^\top w_l)}}. \label{eqn:rho-ij}
\end{align}
Observe that the terms $u_i$ interact with the factor weights $\lambda_{il}$, while the terms $v_i$ interact only with the noise terms $R_l$.

In the rest of this section, we will bound $E_{ij}$ and $\rho_{ij}$ with different choices of $\{w_l\}_{l=1}^L$.

\subsection{Plugin projections}

We now assume we have plugin estimates $(w_l)$ that are close to the inverse factors $(v_l)$: $\|w_l
- v_l\|_2 \le O(\gamma)$ for $l \in [k]$. Then, 
\begin{align*}
w_l^\top u_i &= 
  (v_l + (w_l-v_l))^\top u_i \\
  &= v_l^\top u_i + ||w_l-v_l||_2 \cdot \frac{(w_l-v_l)^\top u_i }{||w_l-v_l||_2} \\
  &= v_l^\top u_i + O(\gamma).
\end{align*}
Recall that $V = U\inv$, so $v_l^\top u_i = \delta_{il}$. 

It will be useful to keep track of $\|\lambda_i\|_2^2$,
\begin{align}
  \|\lambda_i\|^2_2 
  &= \sum_{l=1}^L \pi_i^2 (w_l^\top u_i)^2  \nonumber \\
  &= \pi^2_i \sum_{l=1}^k (v_l^\top u_i + O(\gamma))^2 \nonumber \\
  &= \pi^2_i + O(\gamma) \label{eqn:lambda-no}.
\end{align}

\begin{lemma}[Modulus of uniqueness for plugin projections]
  Let $w_1, \ldots, w_k$ be approximations of $v_1, \ldots, v_k$: $\|w_l - v_l\|_2 \le O(\gamma)$ for $l \in [k]$,
    and let $\sMh = \{\Mh_1, \ldots, \Mh_L\}$ be constructed via
    projection of $\Th$ along $w_1, \ldots, w_L$. Then, for $i \neq j$,
%    Let $w_1, \cdots, w_k$ be approximately equal to $v_1, \cdots, v_k$:
%    $\|w_i - v_i\| \le \gamma \le C\e$ for some constant $C$. 
%   Also assume that the $(v_i)$ are incoherent: $v_i^\top v_j \le \mu$ when $j \neq i$.
  \label{lem:rho-ij-plugin}
  \begin{align*}
    \rho_{ij}^2 
    &\le O(\gamma),
  \end{align*}
%  where $\mu_0 \eqdef \left(\frac{(k-2)\mu + 2}{1 - (k-1)\mu^2}\right)^2$. 
%  Furthermore, for $\mu \le \frac{1}{2d}$, $\mu_0 \le 4$.
\end{lemma}
\begin{proof}
  Let us first bound the numerator of \equationref{rho-ij}.
  \begin{align*}
    ( \lambda_i^\top \lambda_j )^2
    &= \pi_i^2 \pi_j^2 \left( \sum_{l=1}^L w_l^\top u_i u_j^\top w_l \right)^2 \\
    &= \pi_i^2 \pi_j^2 \left( \sum_{l=1}^L v_l^\top u_i u_j^\top v_l + O(\gamma) \right)^2 \\
    &= \pi_i^2 \pi_j^2 \delta_{ij} + O(\gamma) \\
    &= O(\gamma).
  \end{align*}

  Using \equationref{lambda-no}, we get that 
  \begin{align*}
    \rho_{ij}^2 
    &= \frac{O(\gamma)}{(1+O(\gamma))(1+O(\gamma))} \\
    &= O(\gamma).
  \end{align*}
  where in the last line we used the fact that $\frac{1}{1-x} = 1 + x + o(x)$.
\end{proof}

\begin{lemma}[Bound on $E_{ij}$ for non-orthogonal plugin projections]
  \label{lem:e-ij-non-ortho-plugin}
  Let $w_1, \ldots, w_k$ be approximations of $v_1, \ldots, v_k$: $\|w_l - v_l\|_2 \le O(\gamma)$ for $l \in [k]$,
    and let $\sMh = \{\Mh_1, \ldots, \Mh_L\}$ be constructed via
    projection of $\Th$ along $w_1, \ldots, w_L$.
  \begin{align*}
    |E_{ij}| 
    &\le 
      \left(\frac{1}{\pi_i^2} + \frac{1}{\pi_j^2}\right) \|V^\top\|_2~
      p_{ij}^\top r_{ij} + O(\gamma), 
  \end{align*}
  where $p_{ij} \eqdef |\pi_i| \frac{v_i}{\|v_i\|_2} + |\pi_j| \frac{v_j}{\|v_j\|_2}$.
\end{lemma}
\begin{proof}
  Let us bound each term within our expression for $E_{ij}$ (Equation \eqref{eqn:e-ij-non-ortho}).
  \begin{align*}
    \sum_{l=1}^k w_l^\top u_j r_{ij}^\top w_l 
    &= \sum_{l=1}^k v_l^\top u_j r_{ij}^\top v_l + O(\gamma) \\
    &\le r_{ij}^\top v_j + O(\gamma).
  \end{align*}

  Similarly,
  \begin{align*}
    \sum_{l=1}^k w_l^\top u_i r_{ij}^\top w_l 
    &\le r_{ij}^\top v_i + O(\gamma),
  \end{align*}

  From Equation \eqref{eqn:lambda-no}, we have
  \begin{align*}
    \|\lambda_i\|_2^2 &=  |\pi_i|^2 + O(\gamma) \\
    \|\lambda_j\|_2^2 &=  |\pi_j|^2 + O(\gamma).
  \end{align*}

  From \lemmaref{rho-ij-plugin} we have that 
  \begin{align*}
    \rho_{ij}^2 &\leq O(\gamma) \\
    \frac{1}{1 - \rho_{ij}^2} 
    &\leq \frac{1}{1 - O(\gamma)} + O(\gamma) \\
    &\leq 1 + O(\gamma).
  \end{align*}

  Finally,
  \begin{align*}
    |E_{ij}| 
    &\le 
      \left(\frac{1}{\pi_i^2} + \frac{1}{\pi_j^2}\right)
      \left( (|\pi_i| v_i 
              +  |\pi_j| v_j)^\top r_{ij}
              \right) + O(\gamma) \\
    &\le 
      \left(\frac{1}{\pi_i^2} + \frac{1}{\pi_j^2}\right)
      \|V^\top\|_2~ p_{ij}^\top r_{ij}
      + O(\gamma).
  \end{align*}
\end{proof}

Note that the error terms depend not on $u_i$ but rather $v_i$. This is
because the projections $(w_l)$ are chosen to be close to the $v_i$.
Now, let us bound the error in $u_i$.

\begin{theorem}[Non-orthogonal tensor factorization with plug-in projections]
  \label{thm:non-ortho-plugin-1}
  Let $w_1, \ldots, w_k$ be approximations of $v_1, \ldots, v_k$: $\|w_l - v_l\|_2 \le O(\e)$ for $l \in [k]$ and let $\sMh = \{\Mh_1, \ldots, \Mh_L\}$ be constructed via
    projection of $\Th$ along $w_1, \ldots, w_L$.
   Then, for all $j \in [k]$,
 \begin{align*}
    \| \ut_j - u_j \|_2 
    &\le 
    8\e~ \frac{\sqrt{\|\pi\|_1 \pi_{\max}}}{\pi_{\min}^2}  \|V^\top\|_2^3
                + o(\e).
  \end{align*}
\end{theorem}
\begin{proof}
  From \lemmaref{afsari-1} we have that 
  \begin{align*}
    \| \ut_j - u_j \|_2 & \leq \e \sqrt{\sum_{i=1}^d E_{ij}^2} + o(\e),
  \end{align*}
  for $j \in [k]$, where $E_{ij}$ is bounded in \lemmaref{e-ij-non-ortho-plugin} as follows:
  \begin{align*}
    |E_{ij}| 
    &\le 
      \left(\frac{1}{\pi_i^2} + \frac{1}{\pi_j^2}\right)
      \|V^\top\|_2~ p_{ij}^\top r_{ij} + O(\e) \\
  &\le 
  \frac{2}{\pi_{\min}^2}
      \|V^\top\|_2~ p_{ij}^\top r_{ij} + O(\e).
  \end{align*}

  Consequently,
  \begin{align*}
    \| \ut_j - u_j \|_2 
    &\le \e \sqrt{\sum_{i\neq j}^d E_{ij}^2} \\
    &\le 
    \frac{2\e}{\pi_{\min}^2}
    \sqrt{ \sum_{i\neq j}^d \left( 
      \|V^\top\|_2~ p_{ij}^\top r_{ij}
                + O(\e)
             \right)^2
          }
                + o(\e) \\
    &\le 
    \frac{4\e}{\pi_{\min}^2}
    \left( 
    \sqrt{ \sum_{i\neq j}^d 
    \left(
      \|V^\top\|_2~ p_{ij}^\top r_{ij}
    \right)^2 } +
     \right)
                + o(\e),
              \end{align*}
     where we have used the fact that $(a + b)^2 \le 2 (a^2 + b^2)$ and that $\sqrt{a+b} \le \sqrt{a} + \sqrt{b}$.

     From \lemmaref{bounding-pi} we have, $p_{ij}^\top r_{ij} \le 4 \|\pi\|_1 \pi_{\max} \|V^\top\|_2^4$,
  \begin{align*}
    \| \ut_j - u_j \|_2 
    &\le 
    \frac{4\e}{\pi_{\min}^2}
    \left( 
    \sqrt{4 \|\pi\|_1 \pi_{\max} \|V^\top\|_2^6}
     \right)
                + o(\e) \\
    &\le 
    8\e~ \frac{\sqrt{\|\pi\|_1 \pi_{\max}}}{\pi_{\min}^2}  \|V^\top\|_2^3
                + o(\e).
  \end{align*}
\end{proof}

\subsection{Random projections}

We now study the case where the random projections, $(w_l)$, are drawn
  from a standard Gaussian distribution.
First let us show that the modulus of uniqueness $\rho_{ij}$ sharply
  concentrates around $u_i^\top u_j$.
\begin{lemma}[Modulus of Uniqueness with random projections]
  \label{lem:rho-ij}
  Let $w_1, \cdots w_L \in \mathbb R^d$ be entries drawn i.i.d.\ from the standard Normal distribution.
  Let $L > 16 \log(3/\delta)^2$
  Then, with probability at least $1 - \delta$,
  \begin{align*}
    \rho_{ij} 
      &\le 
      u_i^\top u_j + \frac{10 \log(3/\delta)}{\sqrt{L}}.
  \end{align*}
\end{lemma}
\begin{proof}
  Observe from \equationref{rho-ij} that the numerator and the
  denominator of $\rho_{ij}$ are essentially distributed as a $\chi^2$
  distribution (\lemmaref{gaussian-prod}). Thus,
  with probability at least $1-\delta/3$ each, the following hold,
  \begin{align*}
    \frac{1}{L} \sum_{l=1}^L w_l^\top u_i u_j^\top w_l &\le u_i^\top u_j + \|u_i\|_2 \|u_j\|_2 \left(3\sqrt{\frac{\log(3/\delta)}{L}} \right) \\
    \frac{1}{L} \sum_{l=1}^L (w_l^\top u_i)^2 &\ge \|u_i\|_2 \left(1 - \frac{2 \log(3/\delta)}{\sqrt{L}} \right) \\
    \frac{1}{L} \sum_{l=1}^L (w_l^\top u_j)^2 &\ge \|u_j\|_2 \left(1 - \frac{2 \log(3/\delta)}{\sqrt{L}} \right).
  \end{align*}
  Noting that $\|u_i\|_2 = \|u_j\|_2 = 1$ and applying a union bound on
  the above three events, we get that with probability at least
  $1-\delta$,
  \begin{align*}
    \rho_{ij} &\le 
    \frac{u_i^\top u_j + 3\sqrt{\frac{\log(3/\delta)}{L}}}
      {1 - \frac{2 \log(3/\delta)}{\sqrt{L}}}.
  \end{align*}
  Under the conditions on $L$, $\frac{2 \log(3/\delta)}{\sqrt{L}} \le
  \half$. Applying the property that when $x < \half$, $\frac{1}{1-x}
  \le 1+2x$,
  \begin{align*}
    \frac{1}{1 - \frac{2 \log(3/\delta)}{\sqrt{L}}} 
      &\le 1 + \frac{4 \log(3/\delta)}{\sqrt{L}} < 2.
  \end{align*}

  Finally,
  \begin{align*}
    \rho_{ij} &\le 
      \left( u_i^\top u_j + 3\sqrt{\frac{\log(3/\delta)}{L}} \right)
      \left( 1 + \frac{4 \log(3/\delta)}{\sqrt{L}} \right) \\
      &\le 
      u_i^\top u_j \left( 1 + \frac{4 \log(3/\delta)}{\sqrt{L}} \right) 
      + 3\sqrt{\frac{\log(3/\delta)}{L}} \times 2 \\
      &\le 
      u_i^\top u_j + \frac{10 \log(3/\delta)}{\sqrt{L}}.
  \end{align*}
\end{proof}

Let's now bound the inverse modulus of uniqueness.  
\begin{lemma}[Bounding inverse modulus of uniqueness]
  \label{lem:mod-u-inv}
  Let $w_1, \cdots w_L \in \mathbb R^d$ be entries drawn i.i.d.\ from the standard Normal distribution.
  Assume incoherence $\mu$ for that the $(u_i)$: $u_i^\top u_j \le \mu$ for $i\neq j$.
  Let $L_0 \eqdef \left(\frac{50}{(1 - \mu^2)}\right)^2$
  Let $L \ge L_0 \log(3/\delta)^2$.
  Then, with probability at least $1 - \delta$,
  \begin{align*}
    \frac{1}{1 - \rho_{ij}^2} 
    &\le \frac{1}{ 1 - (u_i^\top u_j)^2} \left(1 + \sqrt{\frac{L_0}{L}} \log(3/\delta) \right).
  \end{align*}
\end{lemma}
\begin{proof}
  From \lemmaref{rho-ij}, we have that with probability at least $1 - \delta$,
  \begin{align*}
    \rho_{ij} &\le 
      u_i^\top u_j + \frac{10 \log(3/\delta)}{\sqrt L}.
  \end{align*}
  Then,
  \begin{align*}
    \rho_{ij}^2 
    &\le (u_i^\top u_j)^2 + 2u_i^\top u_j \left( \frac{10 \log(3/\delta)}{\sqrt L} \right) + \left( \frac{10 \log(3/\delta)}{\sqrt L} \right)^2.
  \end{align*}
  Given the assumptions on $L$, we have
  that $L \ge L_0 \log(3/\delta)^2 \ge 50 \log(3/\delta)^2$ and thus
  $\frac{10 \log(3/\delta)}{\sqrt L} \le \half$:
  \begin{align*}
    \rho_{ij}^2 
    &\le (u_i^\top u_j)^2 + 2 \left(\frac{10 \log(3/\delta)}{\sqrt L} \right) + \half \frac{10 \log(3/\delta)}{\sqrt L} \\
    &= (u_i^\top u_j)^2 + \frac{25 \log(3/\delta)}{\sqrt L}.
  \end{align*}
  
  Now, we bound $\frac{1}{1-\rho_{ij}^2}$,
  \begin{align*}
    \frac{1}{1 - \rho_{ij}^2} 
      &\le \frac{1}{ 1 - (u_i^\top u_j)^2 - \frac{25 \log(3/\delta)}{\sqrt L}} \\
      &\le \frac{1}{ 1 - (u_i^\top u_j)^2} \frac{1}{ 1 - \frac{25 \log(3/\delta)}{(1 - (u_i^\top u_j)^2) \sqrt L}} \\
      &\le \frac{1}{ 1 - (u_i^\top u_j)^2} \frac{1}{ 1 - \frac{25 \log(3/\delta)}{(1 - \mu^2) \sqrt L}} \\
      &\le \frac{1}{ 1 - (u_i^\top u_j)^2} \frac{1}{ 1 - \frac{1}{2} \log(3/\delta) \sqrt{\frac{L_0}{L}}}.
  \end{align*}
  Again, given assumptions on $L$, $\frac{1}{2}\log(3/\delta)\sqrt{\frac{L_0}{L}} \le \half$.
  Using the identity that if $x < \half$, $\frac{1}{1-x} \le 1+2x$,
  \begin{align*}
    \frac{1}{1 - \rho_{ij}^2} 
    &\le \frac{1}{ 1 - (u_i^\top u_j)^2} \left(1 + \log(3/\delta)\sqrt{\frac{L_0}{L}} \right).
  \end{align*}
\end{proof}

We are now ready to bound the termwise entries of $E$.
\begin{lemma}[Concentration of $E_{ij}$]
  \label{lem:e-ij-non-ortho}
Let $w_1, \ldots, w_L$ be i.i.d.~random Gaussian vectors $w_l \sim
  \sN(0, I)$, and let $\sMh = \{\Mh_1, \ldots,
  \Mh_L\}$ be constructed via projection of $\Th$ along $w_1, \ldots,
  w_L$.
  Assume incoherence $\mu$ for that the $(u_i)$: $u_i^\top u_j \le \mu$ for $i\neq j$.
Furthermore, let $L \ge L_0 \log(15/\delta)^2$. Then, with probability at least $1 - \delta$,
  \begin{align*}
    |E_{ij}|
    & \le
    \left({\frac{1}{\pi_i^2} + \frac{1}{\pi_j^2}}\right) 
    \left(
    \frac{ \pb_{ij} ^\top r_{ij}}{1 - (u_i^\top u_j)^2} +
    \frac 
    {\pib_{ij} \|r_{ij}\|_2}
      {1 - (u_i^\top u_j)^2}
      \frac{\left( 20 + \sqrt{L_0} \right) \log(15/\delta)}{\sqrt{L}}
    \right),
  \end{align*}
  where $\pb_{ij} \eqdef |\pi_i|u_i + |\pi_j|u_j$ and $\pib_{ij} \eqdef |\pi_i| + |\pi_j|$.
\end{lemma}
\begin{proof}
  Each term in \equationref{e-ij-non-ortho} concentrates sharply about its mean value. We bound each in turn.

  First, consider $\|\lambda_i\|_2^2/L = \frac{1}{L} |\pi_i|^2
  \sum_{l=1}^L (w_l^\top u_i)^2$. With probability at least $1 - \delta
  / 5$ each, the following hold,
  \begin{align*}
    \frac{1}{L} \|\lambda_i\|_2^2 &\ge \pi_i^2 \|u_i\|^2_2 \left( 1 - \frac{2 \log(5/\delta)}{\sqrt L} \right) \\
    \frac{1}{L} \|\lambda_j\|_2^2 &\ge \pi_j^2 \|u_j\|^2_2 \left( 1 - \frac{2 \log(5/\delta)}{\sqrt L} \right).
  \end{align*}
  Thus, using the fact that $\|u_i\|_2^2 = 1$,
  \begin{align*}
    L \left( \frac{1}{\|\lambda_i\|_2^2} +  
    \frac{1}{\|\lambda_j\|_2^2} \right) 
    &\le \frac{\frac{1}{\pi_i^2} + \frac{1}{\pi_j^2}}{1 - \frac{2 \log(5/\delta)}{\sqrt L}}.
  \end{align*}
  Given our assumption on $L$, it follows that $\frac{2 \log(5/\delta)}{\sqrt L} \le \half$. Thus we can use the fact that $\frac{1}{1-x} \le 1 + 2x$ when $x \le \half$ to obtain the following bound:
  \begin{align*}
    L \left( \frac{1}{\|\lambda_i\|_2^2} +  
    \frac{1}{\|\lambda_j\|_2^2} \right)  
    &\le \left({\frac{1}{\pi_i^2} + \frac{1}{\pi_j^2}}\right) \left( 1 + \frac{4 \log(5/\delta)}{\sqrt L} \right).
  \end{align*}

  Next, we bound $\frac{1}{L} \sum_{l=1}^L w_l^\top u_i r_{ij}^\top w_l$ and
  $\frac{1}{L} \sum_{l=1}^L w_l^\top u_j r_{ij}^\top w_l$. From \lemmaref{gaussian-prod}, we have with probability at least
  $1 - \delta/5$ each,
  \begin{align*}
    \frac{1}{L} \sum_{l=1}^L w_l^\top u_j r_{ij}^\top w_l &\le r_{ij}^\top u_j 
      + \|r_{ij}\|_2\|u_{j}\|_2 
      \left(
      3\sqrt{\frac{\log(5/\delta)}{L}}  
      \right) \\
    \frac{1}{L} \sum_{l=1}^L w_l^\top u_i r_{ij}^\top w_l 
    &\le r_{ij}^\top u_i 
      + \|r_{ij}\|_2\|u_{i}\|_2 
      \left(
      3\sqrt{\frac{\log(5/\delta)}{L}}  
      \right).
  \end{align*}
  Note that by definition, $\|u_i\|_2 = 1$.

  Using \lemmaref{mod-u-inv}, we have that with probability at least $1 - \delta/5$,
  \begin{align*}
    \frac{1}{1 - \rho_{ij}^2} 
    &\le \frac{1}{ 1 - (u_i^\top u_j)^2} \left(1 + \sqrt{\frac{L_0}{L}} \log(15/\delta) \right).
  \end{align*}

  Putting it all together, we get that with probability at least $1 - \delta$,
  \begin{align*}
    |E_{ij}|
    &\le 
      \frac{1}{ 1 - (u_i^\top u_j)^2} \left(1 + \sqrt{\frac{L_0}{L}} \log(15/\delta) \right)
      \left({\frac{1}{\pi_i^2} + \frac{1}{\pi_j^2}}\right) 
      \left( 1 + \frac{4 \log(5/\delta)}{\sqrt L} \right) \\
    &\quad
      \left( |\pi_i| r_{ij}^\top u_i + |\pi_j| r_{ij}^\top u_j
      + (|\pi_i| + |\pi_j|) \|r_{ij}\|_2
      \left(
      3\sqrt{\frac{\log(5/\delta)}{L}}  
      \right)
      \right).
    \end{align*}
    Let us define $\pb_{ij} \eqdef |\pi_i|u_i + |\pi_j|u_j$ and $\pib_{ij} \eqdef |\pi_i| + |\pi_j|$:
  \begin{align*}
    |E_{ij}|
    &\le 
      \frac{1}{ 1 - (u_i^\top u_j)^2} \left(1 + \sqrt{\frac{L_0}{L}} \log(15/\delta) \right)
      \left({\frac{1}{\pi_i^2} + \frac{1}{\pi_j^2}}\right) 
      \left( 1 + \frac{4 \log(5/\delta)}{\sqrt L} \right) \\
    &\quad
    \left( \pb_{ij}^\top r_{ij} + \pib_{ij} \|r_{ij}\|_2
      \left(
      3\sqrt{\frac{\log(5/\delta)}{L}}  
      \right)
      \right).
    \end{align*}

  Given that $L \ge L_0 \log(15/\delta)^2$, we have that
  $\sqrt{\frac{L_0}{L}}\log(15/\delta) \le 1$ and $\frac{4
  \log(5/\delta)}{\sqrt L} \le 1$, thus
  \begin{align*}
    \left(1 + \sqrt{\frac{L_0}{L}}\log(15/\delta) \right) \left(1 + \frac{4 \log(5/\delta)}{\sqrt L} \right) 
    &\le 2 \times 2 \\
    &\le 4.
  \end{align*}

  Finally, note that $|\pi_i| r_{ij}^\top u_i + |\pi_j| r_{ij}^\top u_j
  \le (|\pi_i| + |\pi_j|) \|r_{ij}\|_2$, giving us, 
  \begin{align*}
    |E_{ij}| 
    & \le
    \left({\frac{1}{\pi_i^2} + \frac{1}{\pi_j^2}}\right) 
    \frac{\left(\pb_{ij}^\top r_{ij} \right)}{ 1 - (u_i^\top u_j)^2}
     \\
    &\quad +
    \left({\frac{1}{\pi_i^2} + \frac{1}{\pi_j^2}}\right) 
    \frac{\pib_{ij} \|r_{ij}\|_2 }{ 1 - (u_i^\top u_j)^2}
      \left(
      \sqrt{\frac{L_0}{L}} \log(15/\delta) + 2 \frac{4 \log(5/\delta)}{\sqrt L}
        + 4 \left(3\sqrt{\frac{\log(5/\delta)}{L}} \right)
      \right) \\
    & \le
    \left({\frac{1}{\pi_i^2} + \frac{1}{\pi_j^2}}\right) 
    \left(
    \frac{ \pb_{ij} ^\top r_{ij}}{1 - (u_i^\top u_j)^2} +
    \frac 
    {\pib_{ij} \|r_{ij}\|_2}
      {1 - (u_i^\top u_j)^2}
      \frac{\left( 20 + \sqrt{L_0} \right) \log(15/\delta)}{\sqrt{L}}
    \right).
  \end{align*}
\end{proof}

Finally, we bound the error in estimating $u_j$.
\begin{theorem}[Non-orthogonal tensor factorization with random projections]
  \label{thm:non-ortho-random-1}
  Let $w_1, \ldots, w_L$ be i.i.d.~random Gaussian vectors, $w_l \sim
    \sN(0, I)$, and let $\sMh = \{\Mh_1, \ldots,
    \Mh_L\}$ be constructed via projection of $\Th$ along $w_1, \ldots,
    w_L$.
    Assume incoherence $\mu$ for both $(u_i)$ and $(v_i)$: $u_i^\top u_j \le \mu$ and $v_i^\top v_j \le \mu$ for $i \neq j$.
  Let $L_0 \eqdef \left(\frac{50}{1 - \mu^2}\right)^2$.
  Let $L \ge L_0 \log(15 d (k-1)/\delta)^2$. 
  Then, with probability at least $1-\delta$ and for $\e$ small enough,
   \begin{align*}
    \|\ut_j - u_j\|_2 
      &\le \frac{8\e}{1-\mu^2} \frac{\sqrt{\|\pi\|_1 \pi_{\max}}}{\pi_{\min}^2} \|V^\top\|_2^2
      \left( 1 + C(\delta) \sqrt{d} \right),
  \end{align*}
  where $C(\delta) \eqdef \frac{20 + \sqrt{L_0}}{\sqrt{L}} \log(15 (d(k-1))/\delta)$.
\end{theorem}

\begin{proof}
  From \lemmaref{afsari-1} we have that 
  \begin{align*}
    \| \ut_j - u_j \|_2 & \leq \e \sqrt{\sum_{i=1}^d E_{ij}^2} + o(\e),
  \end{align*}
  for $j \in [k]$.
  
  Using \lemmaref{e-ij-non-ortho}, we have that with probability at least $1-\delta/(d(k-1))$,
  \begin{align*}
    |E_{ij}|
    & \le
    \left({\frac{1}{\pi_i^2} + \frac{1}{\pi_j^2}}\right) 
    \left(
    \frac{ \pb_{ij} ^\top r_{ij}}{1 - (u_i^\top u_j)^2} +
    \frac 
    {\pib_{ij} \|r_{ij}\|_2}
      {1 - (u_i^\top u_j)^2}
      \frac{\left( 20 + \sqrt{L_0} \right) \log(15(d(k-1))/\delta)}{\sqrt{L}}
    \right) \\
    & \le
    \left(\frac{2}{\pi_{\min}^2}\right) 
    \left(\frac{1}{1 - \mu^2} \right)
    \left(\pb_{ij} ^\top r_{ij} + 2|\pi_{\min}| \|V^\top\|_2^2 
      \frac{\left( 20 + \sqrt{L_0} \right) \log(15(d(k-1))/\delta)}{\sqrt{L}}
    \right) \\
    & \le
    \left(\frac{2}{\pi_{\min}^2}\right) 
    \left(\frac{1}{1 - \mu^2} \right)
    \left(\pb_{ij} ^\top r_{ij} + 2|\pi_{\min}| \|V^\top\|_2^2 C(\delta) 
    \right),
  \end{align*}
  where we have defined $C(\delta) \eqdef \frac{20 + \sqrt{L_0}}{\sqrt{L}} \log(15(d(k-1))/\delta)$ and are 
  using the fact that $u_i^\top u_j \le \mu$ and $\pib_{ij} = |\pi_i| + |\pi_j| \le 2 |\pi_{\max}|$.

  Applying a union bound on all the entries of $E_{ij}$, we arrive at the following bound for all $j$.
  \begin{align*}
    \|\ut_j - u_j\|_2 
    &\le \e \sqrt{\sum_{i \neq j} E_{ij}^2} \\
    &\le \frac{2\e}{\pi_{\min}^2 (1 - \mu^2)} 
    \sqrt{\left( 
    \sum_{i\neq j}^d \pb_{ij}^\top r_{ij} +
    2 \pi_{\max} \|V^\top \|_2^2 C(\delta)
      \right)^2} \\
    &\le \frac{4\e}{\pi_{\min}^2 (1 - \mu^2)} 
      \left(
      \sqrt{ \sum_{i\neq j}^d (\pb_{ij}^\top r_{ij})^2 } +
      2 \pi_{\max} \|V^\top\|_2^2 C(\delta) \sqrt{ \sum_{i\neq j}^d 1} 
      \right) \\
    &\le \frac{4\e}{\pi_{\min}^2 (1 - \mu^2)} 
      \left(
      \sqrt{ \sum_{i\neq j}^d (\pb_{ij}^\top r_{ij})^2 } +
      2 \pi_{\max} \|V^\top\|_2^2 C(\delta) \sqrt{d}
      \right).
  \end{align*}
  where we use the fact that $(a+b)^2 \le 2(a^2 + b^2)$.

  By \lemmaref{bounding-pi} we also have, $\sum_{j\neq i}^d (\pb_{ij}^\top r_{ij})^2 \le 4 \|\pi\|_1 \pi_{\max} \|V^\top\|_2^4$.
  Finally, note that $\pi_{\max} \le \sqrt{\pi_{\max} \| \pi \|_1}$:
  \begin{align*}
    \|\ut_j - u_j\|_2 
    &\le \frac{4\e}{\pi_{\min}^2 (1 - \mu^2)} 
      \left(
      \sqrt{4 \|\pi\|_1 \pi_{\max}} \|V^\top\|_2^2 +
      2 \pi_{\max} \|V^\top\|_2^2
      C(\delta) \sqrt{d}
      \right) \\
      &\le \frac{8\e}{1-\mu^2} \frac{\sqrt{\|\pi\|_1 \pi_{\max}}}{\pi_{\min}^2} \|V^\top\|_2^2
      \left( 1 + C(\delta) \sqrt{d} \right).
  \end{align*}
\end{proof}

%\subimport{}{extn-proof}
\section{Proofs of auxiliary lemmas} 
\label{app:aux-proofs}

In this section, we prove some auxiliary results that appear as
intermediate steps in the main lemmas above.
%For context, remember that our goal is to find the CP decomposition of a symmetric $p$-th order tenors $T \in \Re^{d \times d \times d \times \cdots}$ from a noisy estimate, $\hat T = T + \epsilon R$, where $R \in \Re^{d \times d \times d \cdots}$ is a $p$-th order tensor with unit operator norm. Recall that we considered $T = \sum_{i=1}^k \pi_i u_i \otimes u_i \otimes u_i \cdots$  where $\{\pi_i\}$ are factor weights and $\{u_i\}$ are the actual vectors.
%
%We reduce higher order tensors with $p > 3$ to a third order tensor by unfolding the additional modes into the third mode, i.e. $T \in \Re^{d \times d \times d^{p-2}}$. Let $d' = d^{p-2}$. The tensor now has the factorization, $T = \sum_{i=1}^k \pi_i u_i \otimes u_i \otimes \underbrace{(u'_i \otimes \cdots)}_{u'_i}$, where $u_i \in \Re^{d'}$. Note that we can also write $R' \in \Re^{d \times d \times d'}$ which continues to have unit operator norm.

\begin{lemma}[Bounding $p_{ij}^\top r_{ij}$]
  \label{lem:bounding-pi}
  Let $p_{ij} \eqdef \pi_i u_i - \pi_j u_j \in \Re^{d}$ and $r_{ij} \eqdef R(v_i, v_j,
  I) \in \Re^{d}$, 
  where $R$ is a tensor with unit operator norm and 
  where $(u_i) \in \Re^{d}$ are unit vectors and $(v_i) \in \Re^{d'}$ form the columns of the matrix $V$ with bounded 2 norm.
  Then, $${\sum_{i \neq j}^{d} (p_{ij}^\top r_{ij})^2} \le 4 \pi_{\max} \|\pi\|_1 \|V\|_2^4.$$
\end{lemma}
\begin{proof}
  Firstly, note that it is trivial to bound the sum as follows, 
  \begin{align*}
    \sum_{i \neq j}^{d} (p_{ij}^\top r_{ij})^2
      &\le \sum_{i \neq j}^d \|p_{ij}\|_2^2 \|r_{ij}\|_2^2 \\
      &\le 4 (d-1) \pi_{\max}^2 \|V\|_2^4,
  \end{align*}
  using the properties that $p_{ij} \eqdef \pi_i u_i - \pi_j u_j$ and that $R$ has unit operator norm and thus
  $\|p_{ij}\|_2 \le 2 \pi_{\max}$ and $\|r_{ij}\|_2 = \| R(v_i, v_j, I)\|_2 \leq \|V\|_2^2$.

  However, we would like a tighter bound with a lower-order dependence
  on $k$. To do so, let us expand $p_{ij}$,
  \begin{align*}
    \sum_{i \neq j}^d  (p_{ij}^\top r_{ij})^2
    &= \sum_{i \neq j}^d ((\pi_i u_i - \pi_j u_j)^\top r_{ij})^2 \\
    &= \sum_{i \neq j}^d (\pi_i R(v_i, v_j, u_i) - \pi_j R(v_i, v_j, u_j))^2 \\
    &= \sum_{i \neq j}^d \pi_j^2 R(v_i, v_j, u_j)^2 + \sum_{i \neq j}^d \pi_i^2 R(v_i, v_j, u_i)^2 - \sum_{i \neq j}^d 2 \pi_i \pi_j R(v_i, v_j, u_i) R(v_i, v_j, u_j).
  \end{align*}

  Using the assumption that $R$ has unit norm, the latter two terms can
  be bounded by $\|\pi\|_2^2 \|V\|_2^4$ and 2 $\pi_j \|\pi\|_1 \|V\|_2^4$ respectively. 

  \providecommand{\tr}{{\tilde r}}

  We now focus on the first term, $\pi_j^2 \sum_{i \neq j}^d R(v_i, v_j,
  u_j)^2$.
  Note that $R(v_i, v_j, u_j) = R(I, v_j, u_j)^\top v_i = {\tr_j}^\top
    v_i$, where $\tr_j \eqdef R(I, v_j, u_j)$ and $\|{\tr}_j\|_2 \le \|V\|_2$ by
    the operator norm condition on $R$.
%  Intuitively, we are summing the projections of $\tr$ along nearly
%  orthogonal vectors, so we expect that $\sum_{i=1}^{d} (\tr_j^\top v_i)^2
%  \approx \|\tr_j\|^2 = 1$. 
  \begin{align*}
    \sum_{i=1}^{d} (\tr_j^\top v_i)^2 
    &= \|V \tr_j\|_2^2 \\
    &\le \|V \|_2^2 \|\tr_j\|_2^2 \\
    &= \|V \|_2^4
  \end{align*}

  Put together, we get that,
  \begin{align*}
    \sum_{i \neq j}^d  (p_{ij}^\top r_{ij})^2
    &\le \pi_j^2 \|V\|_2^4 + \|\pi\|_2^2 \|V\|_2^4 + 2\pi_i \|\pi\|_1 \|V\|_2^4.
  \end{align*}

  Finally, $\pi_i^2 \le \pi_{\max} \|\pi\|_1$ and, by H\"{o}lder's
  inequality, $\|\pi\|_2^2 \le \pi_{\max} \|\pi\|_1$, giving us,
%  \pl{there's a $\mu^2$ in the theorem, $\mu$ here}
  \begin{align*}
    \sum_{i \neq j}^d  (p_{ij}^\top r_{ij})^2
    &\le 4 \pi_{\max} \|\pi\|_1 \|V\|_2^4.
  \end{align*}
\end{proof}

\section{Concentration Inequalities}
\label{sec:concentration-proofs}

In this section, we present several concentration results that are key
to our results. The $\chi^2$ tail bounds presented in
\citet{laurent2000adaptive} play a key role and are reproduced below.

% \begin{lemma}[$\chi^2$ tail inequalities (\citet{laurent2000adaptive})]
%   \label{lem:chi-tail}
%   Let $x_1, \dots, x_n$ be independent $\chi_1^2$ random variables. For
%   any vector $\gamma = (\gamma_1, \dots, \gamma_n) \in \Re^n$ with
%   positive entries, $\gamma \succeq 0$, and any $t > 0$, 
%   \begin{align*}
%     \Pr \left( \sum_{i=1}^n \gamma_i x_i - \|\gamma\|_1 \ge  2 \|\gamma\|_2 \sqrt{t} + 2 \|\gamma\|_\infty t \right) \le e^{-t} \\
%     \Pr \left( \|\gamma\|_1 - \sum_{i=1}^n \gamma_i x_i \ge 2 \|\gamma\|_2 \sqrt{t} \right) \le e^{-t}.
%   \end{align*}
% \end{lemma}
% \begin{proof}
%   See \citet[Lemma 1]{laurent2000adaptive}.
% \end{proof}

\begin{lemma}[$\chi_k^2$ tail inequality]
  \label{lem:chi-k-tail}
  Let $q \sim \chi_k^2$ be distributed as a chi-squared variable with
  $k$ degrees of freedom. Then, for any $t > 0$,
  \begin{align*}
    \Pr( q - k >  2 \sqrt{kt} + 2 t ) \le e^{-t} \\
    \Pr( k - q > 2  \sqrt{kt} ) \le e^{-t}.
  \end{align*}

  Alternatively, we have that with probability at least $1 - \delta$,
  \begin{align}
    q &\ge k \left(1 - \frac{2\log(1/\delta)}{\sqrt{k}} \right). 
  \end{align}
  and similarly, with probability at least $1 - \delta$,
  \begin{align}
    q &\le k \left(1 + 2\sqrt{\frac{\log(1/\delta)}{k}} + \frac{2\log(1/\delta)}{k} \right).
  \end{align}
\end{lemma}
\begin{proof}
  See \citet[Lemma 1]{laurent2000adaptive}.
\end{proof}

\begin{lemma}[Gaussian quadratic forms]
  \label{lem:gaussian-quad}
  Let $x \sim \sN(0,I) \in \Re^d$ be a random Gaussian vector. If $A$ is
  symmetric, $x^\top A x$ is distributed as the sum of $d$ independent
  $\chi^2$ variables, $\sum_{i=1}^d \lambda_{i}(A) \chi^2_1$, where
  $\lambda_i$ are the eigenvalues of $A$.
%   Consequently, $x^\top A x$ has expected value $\Tr(A)$ and is
%   subexponential with parameters $(4 \|A\|_F^2, 4 \|A\|_2)$.
\end{lemma}
\begin{proof}
  Let $A = \sum_{i=1}^d \lambda_i u_i u_i^\top$ be the
  eigendecomposition of $A$.
  Then, $x^\top A x = \sum_{i=1}^{d} \lambda_i \|u_i^\top x_i\|^2$.
  However, $u_i\top x_i$ is distributed as independent $\chi^2_1$
  random variables.
  Thus, $x^\top A x = \sum_{i=1}^{d} \lambda_i \chi_1^2$.
\end{proof}

\begin{lemma}[Gaussian products]
  \label{lem:gaussian-prod}
  Let $x_i \sim \sN(0,I) \in \Re^d$ for $i=1,\ldots,L$ be random Gaussian vectors. Let $L \ge 4 \log(1/\delta)$. Then,
  \begin{enumerate}
    \item $\sum_{i=1}^L (x_i^\top a)^2$ where $a \in \Re^d$ is distributed as $\|a\|_2^2 \chi^2_L$. Consequently, with probability at least $1-\delta$,
      \begin{align*}
        \frac{1}{L} \sum_{i=1}^L (x_i^\top a)^2 
        &\le \|a\|_2^2 \left(1 + 2\sqrt{\frac{\log(1/\delta)}{L}} + \frac{2\log(1/\delta)}{L} \right) \\
        &\le \|a\|_2^2 \left(1 + 3\sqrt{\frac{\log(1/\delta)}{L}} \right) \\
        \frac{1}{L} \sum_{i=1}^L (x_i^\top a)^2 &\ge \|a\|_2^2 \left(1 - \frac{2\log(1/\delta)}{\sqrt{L}} \right).
      \end{align*}
    \item $\sum_{i=1}^L x_i^\top a b^\top x_i$  $a, b \in \Re^d$ and $a
      \neq b$ is sharply concentrated around $a^\top b$: with
      probability at least $1 - \delta$,
      \begin{align*}
        \frac{1}{L} \sum_{i=1}^L x_i^\top a b^\top x_i 
        &\le a^\top b + \|a\|_2 \|b\|_2 \left( 2\sqrt{\frac{\log(1/\delta)}{L}} + \frac{2\log(1/\delta)}{L} \right) \\
        &\le a^\top b + \|a\|_2 \|b\|_2 \left( 3\sqrt{\frac{\log(1/\delta)}{L}} \right).
      \end{align*}
  \end{enumerate}
\end{lemma}
\begin{proof}
  The first part follows directly from \lemmaref{gaussian-quad} and the
  $\chi^2$ tail bound, \lemmaref{chi-k-tail}.

  For the second part, let $A = \frac{a b^\top + b a^\top}{2}$. 
  Note that $x_i^\top a b^\top x_i = x_i^\top A x_i$. 
  Then, by
  \lemmaref{gaussian-quad}, $x_i^\top A x_i = \lambda_1 \chi_1^2
  + \lambda_2 \chi_1^2$, where $\lambda_1$ and $\lambda_2$ are the eigenvalues of $A$. 
  Furthermore, because $A = \frac{a b^\top + b a^\top}{2}$, one of $\lambda_1$ or
  $\lambda_2$ is negative, and the other is positive. Without loss of
  generality, let $\lambda_1 > 0 > \lambda_2$.

  Applying the $\chi^2$ tail bound, \lemmaref{chi-k-tail}, we get that with probability at least $1 - \delta$,
  \begin{align*}
    \lambda_1 \chi_1^2 &\le 
      \lambda_1 (1 + 2\sqrt{ \frac{\log(2/\delta)}{L}} + 2\frac{\log(2/\delta)}{L}) \\
    |\lambda_2| \chi_1^2 &\ge 
      |\lambda_2| (1 - \frac{2\log(2/\delta)}{\sqrt{L}} ).
  \end{align*}
  Applying a union bound, we get,
  \begin{align*}
    \frac{1}{L}\sum_{i=1}^L x_i^\top a b^\top x_i
    &\le \lambda_1 (1 + 2\sqrt{ \frac{\log(2/\delta)}{L}} + 2\frac{\log(2/\delta)}{L}) 
          + \lambda_2 (1 - \frac{2\log(2/\delta)}{\sqrt{L}} ) \\
    &\le (\lambda_1 + \lambda_2) + |\lambda_1| \left(2\sqrt{ \frac{\log(2/\delta)}{L}} + \frac{2\log(2/\delta)}{L} \right) 
          + |\lambda_2| \frac{2\log(2/\delta)}{\sqrt{L}} \\
    &\le (\lambda_1 + \lambda_2) + (|\lambda_1| + |\lambda_2|) \left(2\sqrt{ \frac{\log(2/\delta)}{L}} + \frac{2\log(2/\delta)}{L} \right).
  \end{align*}
  
  Observe that $\lambda_1 + \lambda_2 = \Tr(A) = a^\top b$. Similarly,
  $|\lambda_1| + |\lambda_2| = \|A\|_* = 2 (\half \|a\|_2 \|b\|_2)$.
  Thus, we finally have that with probability at least $1 - \delta$,
  \begin{align*}
    \frac{1}{L} \sum_{i=1}^L x_i^\top a b^\top x_i
      &\le a^\top b + \|a\|_2 \|b\|_2 \left(2\sqrt{ \frac{\log(2/\delta)}{L}} + \frac{2\log(2/\delta)}{L} \right).
  \end{align*}
\end{proof}

\section{Perturbation bounds for joint diagonalization}
\label{sec:joint-diag-bounds}

In this section, we present minor extensions to the perturbation bounds
of \citet{cardoso1994perturbation} and \citet{afsari2008sensitivity} so
that they apply in the low-rank setting.

\paragraph{Notation}
Let $M_l = U \Lambda_l U^\top + \epsilon R_l$ for $l = 1, 2, \dots, L$
  be a set of $d \times d$ matrices to be jointly diagonalized.
$\Lambda_l \in \Re^{k \times k}$ is a diagonal matrix, $R_l \in
  \Re^{d\times d}$ is an arbitrary unit operator norm matrix and
  $\epsilon$ is a scalar.
In the orthogonal setting, $U \in \Re^{d \times k}$ is orthogonal, while
  in the non-orthogonal setting $U \in \Re^{d \times k}$ is an arbitrary
  matrix with unit operator norm.
Let $\lambda_{il} \eqdef (\Lambda_l)_i$ be the $i$-th factor weight of
  matrix $M_l$.
Finally, we say that a set of matrices $\{M_1, \cdots, M_L\}$, $M_l = \sum_{i=1}^d \lambda_{il} u_i v_i^T$ has joint rank $k$ if $\left| \{ i \mid \sum_{l=1}^L |\lambda_{il}| > 0\} \right| = k$.

\begin{lemma}[\citet{cardoso1994perturbation}]\label{lem:cardoso-1}
  Let $M_l = U \La_l U^\top + \e R_l$, $l \in [L]$, be matrices with common factors $U \in \mathbb R^{d \times k}$ and diagonal $\Lambda_l \in \mathbb R^{k \times k}$.
  %Let $M_l = \sum_{i=1}^k \lambda_{il} u_i u_i^\top$, $l=1,2,...,L$ be matrices of joint rank $k$.
  Let $\Ub \in \Re^{d \times d}$ be a full-rank extension of $U$ with columns
  $u_1, u_2, \dots, u_d$ and let $\Ut \in \Re^{d \times d}$ be
  the orthogonal minimizer
  of the joint diagonalization objective $F(\cdot)$. 
  Then, for all $u_j$, $j \in [k]$, there exists a column $\tilde u_j$ of $\Ut$ such that
  \begin{align}
    \|\ut_j - u_j\|_2 \le \e \sqrt{\sum_{i=1}^{d} E_{ij}^2} + o(\e),
  \end{align}
where $E \in \mathbb R^{d \times k}$ is
\begin{align}
  E_{ij} 
  &\eqdef \frac{\sum_{l=1}^L (\lambda_{il} - \lambda_{jl}) u_j^\top R_l u_i}
  {\sum_{l=1}^L (\lambda_{il} - \lambda_{jl})^2}
\end{align}
when $i\neq j$ and $i \leq k$ or $j \leq k$. We define $E_{ij} = 0$ when $i = j$ and $\lambda_{il} = 0$ when $i > k$.
\end{lemma}
\begin{proof}
  See \citet[Proposition 1]{cardoso1994perturbation}. Note that in the
  low rank setting, the entries of $E_{ij}$ (\citet[Equation 15]{cardoso1994perturbation}) where $i, j > k$ are not
  defined, however, these terms only effect the last $d-k$ columns of
  $\Ut$. The bounds for vectors $u_1,...,u_k$ only depend on $E_{ij}$ where $i \in [d]$ and $j \in [k]$, and these are derived in the low-rank setting in the same way as they are derived in the full-rank proof of 
  \citet{cardoso1994perturbation}.
\end{proof}

We now present the corresponding perturbation bounds in \citet{afsari2008sensitivity} to the low rank setting.
\begin{lemma}[\citet{afsari2008sensitivity}]\label{lem:afsari-1}
  Let $M_l = U \La_l U^\top + \e R_l$, $l \in [L]$, be matrices with common factors $U \in \mathbb R^{d \times k}$ and diagonal $\Lambda_l \in \mathbb R^{k \times k}$.
  %Let $M_l = \sum_{i=1}^k \lambda_{il} u_i u_i^\top$, $l=1,2,...,L$ be matrices of joint rank $k$.
% Let $U \in \mathbb R^{d \times k}$ be the matrix of (possibly non-orthogonal) common factors $u_i$, let $V$ be the minimizer
%   of the joint diagonalization objective $F(\cdot)$, and let $\Ut = V\pinv \in \Re^{d \times k}$.
  Let $\Ub \in \Re^{d \times d}$ be a full-rank extension of $U$ with columns
  $u_1, u_2, \dots, u_d$ and let $\Vb = \Ub\inv$, with rows $v_1, v_2, \dots, v_d$.
  Let $\Vt \in \Re^{d \times d}$ be
  the minimizer
  of the joint diagonalization objective $F(\cdot)$ and let $\Ut = \Vt\inv$. 

  Then, for all $u_j$, $j \in [k]$, there exists a column $\tilde u_j$ of $\Ut$ such that
  \begin{align}
    \|\ut_j - u_j\|_2 \le \e \sqrt{\sum_{i=1}^d E_{ij}^2} + o(\e),
  \end{align}
where the entries of $E \in \mathbb R^{d \times k}$ satisfy the equation
  \begin{align*}
    \begin{bmatrix} E_{ij} \\ E_{ji} \end{bmatrix}
      &= 
\frac{-1}{\gamma_{ij}(1-\rho_{ij}^2)}
\begin{bmatrix} \eta_{ij} & -\rho_{ij} \\ -\rho_{ij} & \eta_{ij}^{-1} \end{bmatrix}
  \begin{bmatrix} T_{ij} \\ T_{ji} \end{bmatrix}.
  \end{align*}
when $i \neq j$ and either $i \leq k$ or $j \leq k$. When $i=j$, $E_{ij} = 0$.
The matrix $T$ has zero on-diagonal elements, and is
defined as
\begin{align*}
T_{ij} &= \sum_l v_i^\top R_l v_j \lambda_{jl}, & \mfor 1 \le j \neq i \le d
\end{align*}
and the other parameters are
\begin{align*}
\gamma_{ij} & = \|\lambda_i\|_2 \|\lambda_j\|_2, &
\eta_{ij} & = \frac{\|\lambda_i\|_2}{\|\lambda_j\|_2}, &
\rho_{ij} & = \frac{\lambda_i^\top \lambda_j}{\|\lambda_j\|_2 \|\lambda_i\|_2}, &
(\lambda_i)_k & = \lambda_{ik}.
\end{align*}
We define $\lambda_{il} = 0$ when $i > k$.
\end{lemma}

\begin{proof}
 In \citet[Theorem
  3]{afsari2008sensitivity} it is shown that $\Vt = (I + \epsilon E)
  V + o(\epsilon)$, where $E_{ij}$ is defined for $i, j \in [d]$ (\citet[Equation 36]{afsari2008sensitivity}). Then,
  \begin{align*}
    \Ut &= \Ut (I + \epsilon E)\inv + o(\epsilon) \\
      &= \Ut (I - \epsilon E) + o(\epsilon).
  \end{align*}
  Note that, once again, in the low rank setting, the entries of
  $E_{ij}$ when $i, j > k$ are not characterized by Afsari's results; however, these terms only
  effect the last $d-k$ columns of $\Ut$.
\end{proof}

\begin{lemma}
\label{lem:no-eij-bound-1}
  Let $M_l = U \La_l U^\top + \e R_l$, $l \in [L]$, be matrices with common factors $U \in \mathbb R^{d \times k}$ and diagonal $\Lambda_l \in \mathbb R^{k \times k}$.
  %Let $M_l = \sum_{i=1}^k \lambda_{il} u_i u_i^\top$, $l=1,2,...,L$ be matrices of joint rank $k$.
% Let $U \in \mathbb R^{d \times k}$ be the matrix of (possibly non-orthogonal) common factors $u_i$, let $V$ be the minimizer
%   of the joint diagonalization objective $F(\cdot)$, and let $\Ut = V\pinv \in \Re^{d \times k}$.
  Let $\Ub \in \Re^{d \times d}$ be a full-rank extension of $U$ with columns
  $u_1, u_2, \dots, u_d$ and let $\Vb = \Ub\inv$, with rows $v_1, v_2, \dots, v_d$.
  Let $\Vt \in \Re^{d \times d}$ be
  the minimizer
  of the joint diagonalization objective $F(\cdot)$ and let $\Ut = \Vt\inv$. 

  Then, for all $u_j$, $j \in [k]$, there exists a column $\tilde u_j$ of $\Ut$ such that
  \begin{align}
    \|\ut_j - u_j\|_2 \le \e \sqrt{\sum_{i=1}^d E_{ij}^2} + o(\e),
  \end{align}
where the entries of $E \in \mathbb R^{d \times k}$ are bounded by
\begin{align*}
  |E_{ij}|
  &\le \frac{1}{1 - \rho_{ij}^2} 
  \left(\frac{1}{\|\lambda_i\|^2_2} + \frac{1}{\|\lambda_j\|^2_2}\right)
  \left( \left|\sum_{l=1}^L v_i^\top R_l v_j \lambda_{jl} \right| + \left|\sum_{l=1}^L v_i^\top R_l v_j \lambda_{il} \right| \right),
\end{align*}
when $i \neq j$ and $E_{ij} = 0$ when $i = j$  and $\lambda_{il} = 0$ when $i > k$.
Here $\lambda_i = (\lambda_{i1}, \lambda_{i2}, ..., \lambda_{iL}) \in \Re^L$ and
$\rho_{ij} = \frac{\lambda_{i}^\top \lambda_{j}}{\|\lambda_{i}\|_2
\|\lambda_{j}\|_2}$ is the modulus of uniqueness, a measure of
how ill-conditioned the problem is.
\end{lemma}
\begin{proof}

From \lemmaref{afsari-1}, we have that
\begin{align*}
\left\| \begin{bmatrix} E_{ij} \\ E_{ji} \end{bmatrix} \right\|
&\leq \frac{\eta_{ij} + \eta_{ji}}{\gamma_{ji}(1-\rho^2_{ij})} 
    \left \| \begin{bmatrix} T_{ij} \\  T_{ji} \end{bmatrix} \right \|,
\end{align*}
where
\begin{align*}
\gamma_{ij} & = \|\lambda_i\|_2 \|\lambda_j\|_2, &
\eta_{ij} & = \frac{\|\lambda_i\|_2}{\|\lambda_j\|_2}, &
\rho_{ij} & = \frac{\lambda_i^\top \lambda_j}{\|\lambda_j\|_2 \|\lambda_i\|_2},
\end{align*}
and the matrix $T$ is defined to be zero on the diagonal and for $i \neq j$
defined as
\begin{align*}
T_{ij} &= \sum_{l=1}^L v_i^\top R_l v_j \lambda_{jl}, & \mfor 1 \le j \neq i \le d
\end{align*}

Taking $\| \cdot \|$ to be the $l_1$-norm in the above expression, we have that
$$ |E_{ij}| \leq |E_{ij}| + |E_{ji}| \leq \frac{\eta_{ij} + \eta_{ji}}{\gamma_{ji}(1-\rho^2_{ij})} \left( |T_{ij}| + |T_{ji}| \right). $$
Since 
$$ \frac{\eta_{ij} + \eta_{ji}}{\gamma_{ji}} = \frac{\|\lambda_i\|_2^2 + \|\lambda_j\|_2^2} {\|\lambda_i\|_2^2 \|\lambda_j\|_2^2} = \frac{1}{\|\lambda_i\|^2_2} + \frac{1}{\|\lambda_j\|^2_2}$$
and
$$ T_{ij} = \sum_{l=1}^L v_i^\top R_l v_j \lambda_{jl}, $$
the claim follows.
\end{proof}

\end{document}